\documentclass[10pt,journal,compsoc]{IEEEtran}

\usepackage{cite}
\usepackage{CJKutf8}
\usepackage{url}
\usepackage{subfigure}
\usepackage{graphicx}
\graphicspath{{eps/}}
\DeclareGraphicsExtensions{.eps}
\usepackage{algorithmic}
\usepackage{algorithm}
\usepackage{amsmath,amssymb,amsfonts}
\usepackage{makecell}
\usepackage{multirow}
\usepackage{amsthm}
\usepackage{threeparttable}

\usepackage{xcolor}
\usepackage{colortbl}
\usepackage{hyperref} 

\begin{document}
\title{Requirements Elicitation in Cognitive Service for Conversational Recommendation}

\author{Bolin~Zhang,
        Zhiying~Tu,~\IEEEmembership{Member,~IEEE,}
        Yunzhe~Xu,
        Dianhui~Chu,~\IEEEmembership{Member,~IEEE,}
        and~Xiaofei~Xu,~\IEEEmembership{Member,~IEEE,}
\IEEEcompsocitemizethanks{\IEEEcompsocthanksitem All authors are with the School of Computer Science and Technology, Harbin Institute of Technology, Weihai 264209 , China. \protect\\
E-mail: \{brolin, tzy\_hit, 181310122, chudh, xiaofei\}@hit.edu.cn
% \IEEEcompsocthanksitem J. Doe and J. Doe are with Anonymous University.
}% <-this % stops an unwanted space
\thanks{This Manuscript based partly on work\cite{agr} presented at conference (ICWS 2021).}

}
% \markboth{IEEE Transactions on Services Computing,~Vol.~14, No.~8, January~2022}%
% {Zhang \MakeLowercase{\textit{et al.}}: Requirements Elicitation in Cognitive Service for Recommendation}

\markboth{}%
{Zhang \MakeLowercase{\textit{et al.}}: Requirements Elicitation in Cognitive Service for Recommendation}

\IEEEtitleabstractindextext{%
\begin{abstract}
Nowadays, cognitive service provides more interactive way to understand users' requirements via human-machine conversation. In other words, it has to capture users' requirements from their utterance and respond them with the relevant and suitable service resources. To this end, two phases must be applied: I.Sequence planning and Real-time detection of user requirement, II.Service resource selection and Response generation. The existing works ignore the potential connection between these two phases. To model their connection, Two-Phase Requirement Elicitation Method is proposed. For the phase I, this paper proposes a user requirement elicitation framework (URef) to plan a potential requirement sequence grounded on user profile and personal knowledge base before the conversation. In addition, it can also predict user's true requirement and judge whether the requirement is completed based on the user's utterance during the conversation. For the phase II, this paper proposes a response generation model based on attention, SaRSNet. It can select the appropriate resource (i.e. knowledge triple) in line with the requirement predicted by URef, and then generates a suitable response for recommendation. The experimental results on the open dataset \emph{DuRecDial} have been significantly improved compared to the baseline, which proves the effectiveness of the proposed methods.
\end{abstract}

}

\maketitle

\IEEEdisplaynontitleabstractindextext
\IEEEpeerreviewmaketitle

\section{Introduction}
\IEEEPARstart{C}{ognitive} conversation draws growing attention in both academia and industry. Since natural languages act as a powerful information carrier, conversation is a natural solution to the long-standing information asymmetry problem in service matching\cite{hci}. Therefore, cognitive conversation can be viewed as a smart service deliverer in many service seeking topics such as searching, recommendation, and question answering. As seen through widely adopted examples such as Google Assistant, Apple Siri, and Amazon Alexa, these new-generation smart service delivery bots, in the era of artificial intelligence, are expected to make it much easier and more direct for users to obtain services via cognitive conversation. By sensing intentions and demands of the user, \cite{esbot} deem those kind of bots are able to deliver appropriate services to the user smartly. However, how to elicit users' requirements naturally and proactively need to be coped with. Given the user requirement, it's hard to select a corresponding service resource and generate the response to recommend this item to user. 

To this end, the Requirements Elicitation in Cognitive Service subjects to the following challenges:
\begin{itemize}
\item[-] \textbf{Sequential requirements}:
    Usually, user's requirements are sequentially correlated. It means that certain requirements often appear together in a specific order. The existing works focus on single requirements detection without considering generate the potential requirement sequence\footnote{A path containing multiple requirements, the nodes have a continuation relationship, and the length is moderate between 3-6.}. Sometimes, user's requirements may be fuzzy and need to be lead into explicit requirements in a potential sequence of requirements. Moreover, without the sequence , bots can not crawl the required service resources in advance, which to meet user requirements.
\item[-] \textbf{Dynamic requirements}:
    During the conversation, user's requirement will change at any time and need be satisfied in multi-turn. Therefore, whether the requirement of the last turn are met or not will affect the creation of a new requirement. The existing works\cite{dynamic} only focus on detecting the current requirement, but lacked judgment on the completion of the requirement. Moreover, after the requirement changes, the potential sequence should also change dynamically in real time be re-planned.
\item[-] \textbf{Appropriate reply}:
    Either in requirement elicitation or response generation, a piece of reply text should be output by generative models. However, the responses generated by this kind of models are usually safe, anomaly, and hollow \footnote{Safe means the bot often responds with universal answers such as "thank you"; anomaly means the bot responds with opposing, false and insulting remarks; hollow means the bot responds with vague and unspecific content due to lack of knowledge of the user.}, so it is difficult to deploy for recommendation\cite{dialoguesurvey}. Thus, the generative models should has the ability to select the relevant service resource and generate an appropriate reply to the user. 
\end{itemize}
% 在对话中， 用户表达他们的意图时，往往是动态变化的一个序列
% 用户表达的特点：意图离散、
% 挑战1 对话需要序列逻辑
% 2 序列是动态变化的
% 3 生成恰当回复

\begin{figure*}[!t]
	\centering
	\includegraphics[scale=0.8]{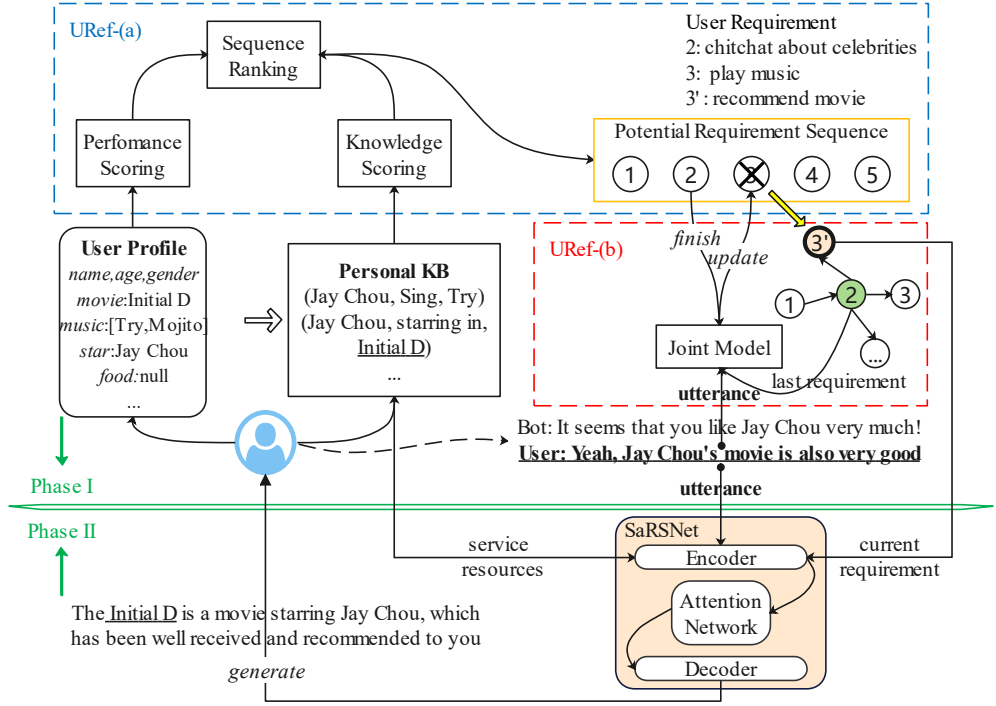}
	\caption{
	The Overview of the proposed Two-phase Requirement Elicitation Method. Above the green line, Phase I is applied: The Requirement Sequence planning is shown in the blue-dotted rectangle and Real-time detection of user requirement is shown in the red-dotted rectangle.
	Below the green line, Phase II is applied: Given the current requirement, user's utterance and service resources, SaRSNet generates an appropriate response to meet user's requirement.
	}
	\label{framework_figure}
\end{figure*}

To address these challenges, two phases need to be applied in succession. In Phase I, the mission of User Requirement Elicitation is divided into two parts: Sequence Planning and Real-time Detection. In Phase II, the Recommend Response Generation task is also divided into two parts: Service Resource Selection and Response Generation.

% \noindent
\textbf{Requirement Elicitation}: firstly, to plan a potential requirement sequence for multi-turn dialogue, and then to detect the user's requirement from their utterance and judge whether requirement be met at the same time. Grounded on user’s preference and personal Knowledge Base (KB), Sequence Planning prepares user's potential requirement chain before the conversation. Based on users' utterances, Real-time Detection predicts user's current-turn requirement during the conversation. By this way, the new requirement is regarded as the starting vertex then re-plans the potential chain when users change their mind. During the conversation, the timing to lead the next potential requirement depends on whether the current one has been fully met. Thus, Real-time Detection is divided into two potentially related sub-tasks: Requirement Completion Estimation and Current Requirement Prediction. The former one is binary-class classification task for predicting whether the current requirement is fully finished. The latter one is a multi-class classification task for predicting the user's current requirement. These two sub-tasks can make full use of the information learned by each other and improve their own performance.    

To this aim, the proposed \textbf{u}ser \textbf{r}equirement \textbf{e}licitation \textbf{f}ramework (\textbf{URef}) with two parts, URef-(a) and URef-(b), will be applied in Phase I (as shown in Fig.~\ref{framework_figure}). URef-(a) will use the domain knowledge, such as music, celebrity, movie, and etc., as the constraints of the requirement sequence planning, due to the lack of domain-related knowledge in the personal KB. For instance, if a user has the most preferred entities in music domain, then URef-(a) will set the sequence to "daily greetings" $\rightarrow$ "chitchat about celebrities" $\rightarrow$ "play music"$\rightarrow$... "goodbye"]. URef-(b) is reponsable to predict whether the current task is end, and decide whether to follow the pre-define planning. For example, during the conversation, URef-(b) realizes that it is the time to kick-off the task node "chitchat about celebrities", Meanwhile, it finds that the user's intent (the current requirement) has become "recommend movie, according to the user's utterance. Thus, it asks URef-(a) to re-plan the sequence with "recommend movie" as a new starting point.

% \noindent
\textbf{Response Generation}: generating an appropriate reply text based on the user's utterance to elicit his/her explicit requirement or satisfy his/her requirement directly. In the personal KB, each SPO (Subject-Predicate-Object) triplet is labeled with a requirement domain, and a triple can be regarded as a service resource. The domains of the service resources is consistent with the domains of the user profile. After elicit user's requirement by URef, Service Resource Selection aims to pick out one of the service resources that belongs to the same domain as user requirements from the personal KB. This allows the bot to meet the user's current requirement. Given users' requirements, user's utterance and the selected service resource, Response Generation aims to output a smooth, appropriate and informative text to response the user.

For this purpose, the proposed recommend response generation model (SaRSNet) will be applied in Phase II. As shown in Fig.~\ref{framework_figure}, \textbf{SaRSNet} is an encoder-decoder \textbf{net}work that \textbf{s}elects the \textbf{a}ppropriate \textbf{r}e\textbf{s}ource (i.e. SPO Triple) in line with requirement of utterance and then generates a suitable response to user. For example, the requirement node ("recommend movie") predicted by URef, then the user's utterance and the service resources will all be taken as input by SaRSNet. Afterwards, one of the triples about the movie will be selected based on attention network and a text containing the service resource (Jay Chou, starring in, Initial D) will be generated as output to response the user. 

The main contributions are summarized as follow:
\begin{enumerate} 
	\item [1)] This paper propose a novel user requirement elicitation framework, URef. Based on user’s preference and personal KB, URef-(a) prepares user's potential requirement sequence. Based on user's utterance, URef(b) judges whether the current requirement have been met and then detect the true requirement.
	\item [2)] This paper propose a response generation model after user requirement elicitation, SaRSNet. Given the user's requirement, SaRSNet Choose the most relevant service resources in personal KB and then generation an appropriate response to user.
	\item [3)] Both URef and SaRSNet conducted experiments on the public dataset \emph{DuRecDial} and the results have been significantly improved compared to the baseline.
\end{enumerate}

The rest of this paper is organized as follows. 
Section \ref{Related Works} summarizes the related works. URef is described in Section \ref{ure} and SaRSNet is described in Section \ref{rrg}. Section \ref{Experiments} shows the experimental setup and Section \ref{Results} analyzes the experimental result. In the end, \ref{conclusion} concludes this paper.

\section{Related Works}\label{Related Works}
The related works are in line with three major research topics: Requirement Elicitation, Text Classification and Response Generation.

\subsection{Requirement Elicitation}
Basically, Requirement Elicitation (RE) is the practice of researching and discovering the requirements of a system from users, customers, and other stakeholders\cite{re21} in software requirements engineering. In cognitive service, RE refers to the process of detecting and understanding user requirements from the original user requirement expression (i.e. utterance). The existing works\cite{intent1,intent2, intent3} focus on interpreting user's requirement by extracting user's intention from their one-off questions. However, only through multi-turn dialogue, the explicit requirement of user can be elicited and satisfied. Due to user's goal-oriented behaviour in a conversation, goal (i.e user requirement) planning has attracted lots of research interests\cite{goalsurvey}. According to a certain requirement of the user, the bot will have a goal in line with it. Endowing it with the ability of proactively leading the dialogue with an explicit goal, Requirement Planning takes a radical step towards building type of human-like conversational agents\cite{FrameworkforGoalDriven}. Although some research works \cite{gp2019,Proactive2019, gp2020} provide more controllability to RE by planning goal, these models can just 
output a single requirement, instead of a requirement sequence. The work \cite{aaai2020} considers the problem of goal planning grounded on knowledge graph (KG). However, this work regards the goal as an entity in KG, actually solves the problem of topic selection in chitchat. In contrast, this paper consider more diverse requirements (e.g. order news, play music, recommend movie), not only limited to chitchat. Furthermore, the existing works only predict the current requirement based on user utterances while this paper plan a potential and reasonable requirement sequence before the conversation.

\subsection{Text Classification}
Text classification (TC) is the task of assigning a sentence or document (e.g. articles, comments, utterances) to an appropriate category\cite{surveyTextClassify}. With the development of deep learning and pre-training models, TC has not only made great progress in traditional tasks such as sentiment analysis\cite{TC2020,TC2021} and news classification\cite{news2020}, but also achieved good results in multi-task learning (e.g. Intent Detection and Slot Filling\cite{IDSF2020,IDSF2021}, Entity and Relation Extraction\cite{ERE2019,ERE2020}, Sentiment and Act Classification\cite{2020DCR,2021DCR}). In this paper, the detection of user requirement can also be regarded as a TC task. The traditional TC task only considers the semantic information of the text, while the joint model is used in multi-task learning framework to realize the transfer and sharing of information between different sub-tasks. Each sub-task can make full use of the information learned by the other's task to improve its own effect\cite{MTLSurvey}. However, these works do not integrate features outside of the semantics and the data labels are static, which cannot effectively model the dependence\footnote{The choice of the next requirement depends to a certain extent on the current requirement. For example, if the current requirement is ``recommend music", the next requirement is likely to be ``play music".} of each requirement in the time series during the conversation. The data label of requirement detection has an order-dependent relationship. By introducing the sub-task of Requirement Completion Estimation, the requirement of the last time step is encoded to deal with the classification task in dynamic scenarios. 

\subsection{Response Generation}
Given the user's utterance, Response Generation (RG) is aiming to generate a consistent and engaging response text\cite{nrm}. RG is an key component of the dialogue system that affects the naturalness of conversation and the user experience\cite{rg}. Inspired by the field of machine translation, the current mainstream approach to response generation is to treat the task as a “translation” of user expressions to machine responses. Since no external knowledge (service resource) is introduced, the responses generated by the end-to-end models proposed in the existing works (\cite{pwork1, pwork2, pwork3}) are relatively hollow and contain less information. Although taking knowledge as additional input can alleviate the problem of generic responses in dialogue systems, choosing right knowledge facts for the current scenarios is still a challenge. To cope with it, the industry has proposed several transformer-based pre-training generative models: Plato2\cite{Plato2} from Baidu with 1.6 billion parameters,  Meenea\cite{Meena} from Google with 2.6 billion parameters, Blender\cite{Blender} from Facebook with 9.4 billion parameters, etc. Using a a large-scale corpus as training data, these models will learn some human response patterns under different dialogue objectives, and the various knowledge contained in the corpus will be implicitly retained by the large number of parameters. But this will take up a lot of computing resources and training time, which is unbearable in some cases. In this way, the user's fine-grained requirements such as ordering restaurants and checking the weather cannot be dealt with in task-oriented dialogues. In cognitive service, when a user puts forward a requirement, the bot needs to inform the user of the service resources (knowledge) which satisfy the requirement\cite{Cognitive}. Therefore, user's requirement (dialogue goal), service resources should be unified as external information as the input of the generative model.

\section{User Requirement Elicitation}
\label{ure}

\subsection{Overview of The Framework}
The whole framework is divided into two parts: URef-(a) and URef-(b). It respectively plans the potential requirement sequence before the conversation and detects the requirements during the conversation. URef-(a) takes User Profile and Personal KB as input, and the output is a requirement sequence that conforms to the user's preference. URef-(a) includes three steps:
\begin{itemize}
    \item[-] \textbf{Performance Scoring}: Calculate the user's preference for a single domain
    \item[-] \textbf{Knowledge Scoring}: Calculate the knowledge richness of triplets in a single domain
    \item[-] \textbf{Sequence Ranking}: According to the calculation of the above modules and the heuristic information in the historical data, the candidate sequence set is screened to find the optimal one.
\end{itemize}

\begin{figure}[htb]
	\centering
	\includegraphics[width=1.0\linewidth]{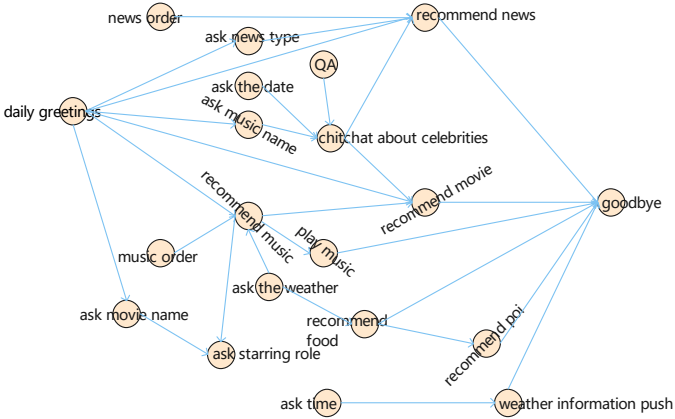}
	\caption{Requirement Transition Backbone Graph. Each node on the graph represents a user requirement. The edges between nodes represent the requirement transition relationship, which is mined from the requirement sequence of users in the dataset. When there is a new kind of requirement appears or no training set, this graph can be manually designed and has a certain scalability.}
	\label{backbone_figure}
\end{figure}

The heuristic information, as shown in Fig.~\ref{backbone_figure}, is a user requirement transition backbone graph constructed according to the sequence of different users in training set. The candidate sequence set is obtained by traversing the backbone graph.

During the conversation, URef-(b) uses completion subnet and prediction subnet to cope with Requirement Completion Estimation and Current Requirement Prediction respectively, and enhances the performance of these two networks by joint learning. The embedding vector of the last utterance's requirement node in the backbone graph and the semantic vector of the current utterance are used as input for URef-(b), which takes the result of requirement completion (binary labels) and requirement prediction (multiple labels) as output.

\renewcommand{\algorithmicrequire}{\textbf{Input:}}
\renewcommand{\algorithmicensure}{\textbf{Output:}}
\begin{algorithm}[htb]
	\caption{Requirement Transition Graph Building Algorithm}
	\label{al1}
	\begin{algorithmic}[1]
		\REQUIRE Historical requirement sequences set $P$,the $k$-th user's requirement sequence is represented by $p_k$, $p_k \in P$;
		$M$, a mapping dictionary for each requirement in $p_k$
		\ENSURE $A[n][n]$, $n=|V|$, the adjacency matrix of Transition Graph $G=(V,E)$
		\STATE initialize matrix $A$ with zero elements
    		\FOR{ $ p_k $ in sequences set $P$ }
    		\STATE len = length($ p_k $)
    		\FOR{$l=0$; $l<len$; $l++$}
    		\STATE $i,j = M(p_k[l]), M(p_k[l+1])$
    		\STATE $A[i][j]++$
    		\ENDFOR
    	\ENDFOR
		\RETURN $A[n][n]$
	\end{algorithmic}
\end{algorithm}

\subsection{Sequence Planning}
The potential requirement sequence can be represented by a graph structure, where each vertex represent a kind of requirement and edge represent two different requirements appear one after another. Before the sequence planning, URef-(a) firstly construct Backbone Graph $G=(V,E)$ by Algorithm \ref{al1}.

\begin{figure*}[htb]
	\centering
	\includegraphics[width=0.8\linewidth]{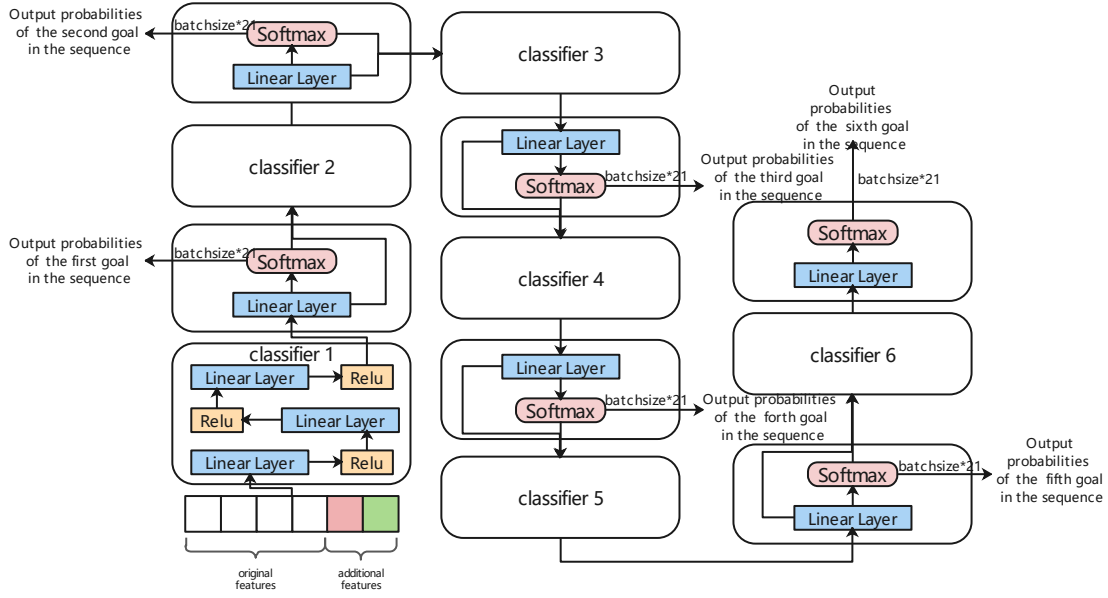}
	\caption{The architecture of SP-MLP. SP-MLP contains 6 classifiers, each of them predict the requirement sequentially.}
	\label{mlp}
\end{figure*}

$V$ is a set of nodes, and $v_i \in V$ denotes a kind of requirement. $E$ is the set of edges, and $e=(v_i, v_j)\in E$ denotes the transition relationship from $v_i$ to $v_j$. The output of Algorithm\ref{al1} is an adjacency matrix of $G$, which is visualized as the network as shown in Fig.~\ref{backbone_figure}.

Then, this paper designs two pipeline strategies to filter the candidate paths in $G$. Strategy 1 first uses the \textit{preference satisfaction} to filter out $top_k$ paths, and finally uses \textit{knowledge abundance} of user's preferred domain to filter out the only path as requirement sequence. Strategy 2 uses \textit{knowledge abundance} first and then \textit{preference satisfaction} for selection. The inspiration of the proposed strategies comes from:

\begin{itemize}
\item[-] A user’s preference for different domains is different. The requirements that are relevant to the user's favorite domain (i.e. the domain contains the most interest entities in profile) should be prioritized.
\item[-] Under the condition of knowledge limited, the domain with the greatest knowledge richness should be prioritized.
\item[-] Strategy 1 and 2 respectively represent two different dialogue modes preference-driven and knowledge-driven.
\end{itemize}

$\{D\}$ denotes a set of types of preferred domains appearing in user profile, the amount of domains is $|D|$. $ \{ent_j^{i}\}$ denotes a list of interest entities of the $i$-th user in the $j$-th domain, and $|ent_j^{i}|$ denotes the amount of entities in it. The set $\{R\}$ contains all user requirements, and $\{D_{r_k}\} \subsetneqq \{D\}$ denotes a set of preferred domains involved in a single requirement $r_k$, $r_k \in \{R\}$. $sat_{r_k}$ denotes the $i$-th user's \textit{preference satisfaction} of $r_k$, formulated as:

\begin{equation}
 sat_{r_k} = \frac{ \sum_{m=1}^{|D_{r_k}|} |ent_m|^{i} }
%  分母
 { \sum_{n=1}^{|D|} |ent_n|^{i} } \in [0,1]
\label{satisfaction}
\end{equation}

$\{kg_i\}$ represents the $i$-th user's personal KB linked with his profile. $t_j$ represents the $j$-th SPO triplet, and $t_j \in \{kg_i\}$, $1<j<|kg_i|$. $\{T_{g_k}^{d_l}\}$ represents a set of triplets of requirement $r_k$ in $l$-th preferred domains, and $k \in [1,|R|]$, $l\in [1,|D|]$. $abd_{r_k}$ denotes \textit{knowledge abundance} of $r_k$, formulated as:

\begin{equation}
 abd_{r_k} = \frac{ \sum_{m=1}^{|D|} |T_{r_k}^{d_m}| }
%  分母
{|kg_i| } \in [0,1] 
\label{abundance}
\end{equation}

The procedure of sequence planning is shown in Algorithm \ref{al2}. Firstly, URef traverse backbone graph built by Algorithm \ref{al1} to get the candidate set of sequences, $ans\_path$. Then Performance and Knowledge Scoring modules will be used to calculate $sat_{r_k}$ and $ abd_{r_k}$, respectively. Finally, all the sequences in $ans\_path$ are scored according to the given strategy, and the one with the highest score is the output.

To compare with Algorithm \ref{al2}, this paper designs a simple Multi Layer Perceptrons Model (SP-MLP) for Sequence Planning, as shown in Fig.~\ref{mlp}. It contains six 21-class classifiers, each of them consists of 3 linear layers and uses $ReLU$ as the activation function to sequentially predict the requirements. Each classifier is connected by a fully connected layer, and use $softmax$ to calculate the probability distribution of the 21 labels (20 requirement-labels and one none-label) at its position. The $cross\_entropy$ is adopted as the loss function of SP-MLP. Apart from the features already given in the dataset, SP-MLP takes user preferences and knowledge abundance as the additional input. 

\renewcommand{\algorithmicrequire}{\textbf{Input:}}
\renewcommand{\algorithmicensure}{\textbf{Output:}}
\begin{algorithm}[!ht]
	\caption{Requirement Sequence Planning Procedure}
	\label{al2}
	\begin{algorithmic}[1]
		\REQUIRE Backbone graph $G$; $i$-th user's profile $up_i$; , $i$-th user's personal KB $kg_i$; strategy identifier $s$, picked out from 1 and 2; $top_k$, threshold of path numbers in the first filter;  
		\ENSURE $path$, a sensible sequence of requirement nodes
		\STATE initialize satisfaction and abundance score list $S$, $A$
		\STATE initialize candidate path set C
		\STATE $P=Traverse(G, node=None)$ \COMMENT{if $node$ specified, traverse all the paths starting from the given requirement in $G$ }
		\FOR{ $p$ in path set $P$ }
		\STATE initialize satisfaction and abundance score of $p$, $s_p$, $a_p$
    		\FOR{ $r$ in path $p$}
    		\STATE $s_p = sat_r ++$
            \STATE $a_p = abd_r ++$
            \ENDFOR
            \STATE $S.append(s_p)$
            \STATE $A.append(a_p)$
		\ENDFOR
		\IF{s==1}
		\STATE $C=max(S, top_k)$ \COMMENT{find the largest $top_k$ paths in $S$}
		\ELSE
		\STATE $C=max(A, top_k)$ 
		\ENDIF
		\STATE $path=max(C, 1)$ 
		\RETURN $path$
	\end{algorithmic}
\end{algorithm}

\subsection{Real-time Detection}

\begin{figure*}[htb]
	\centering
	\includegraphics[width=0.8\linewidth]{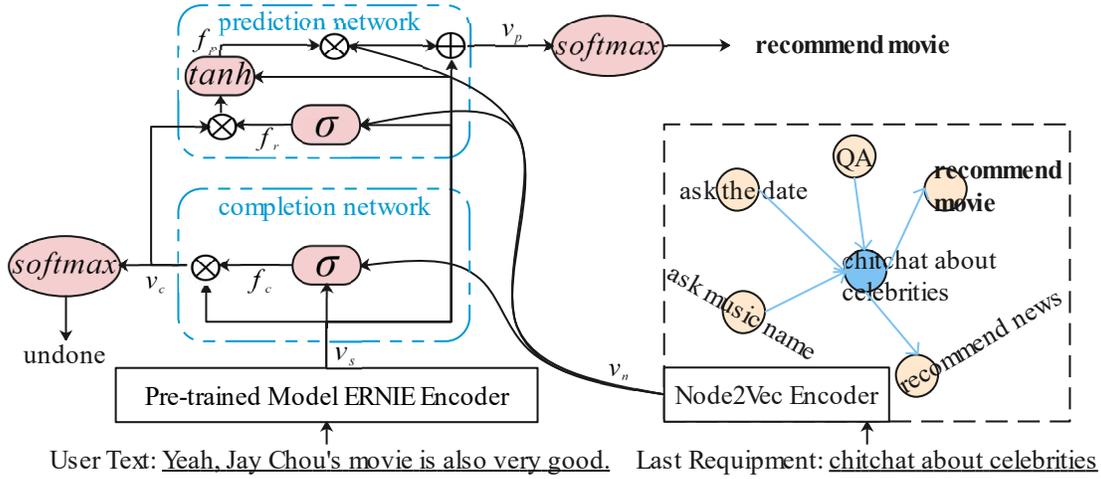}
	\caption{The Architecture of Joint Model in URef-(b), including two encoders and two novel networks}
	\label{jointmodel}
\end{figure*}

The architecture of the joint model in URef-(b) is shown in Fig.~\ref{jointmodel}, contains three modules: Encoder, Completion Network and Completion Network. The model takes the current utterance and the last utterance's requirement as the input. $u_t$ represents the utterance of one user at time $t$, and utterance's requirement at the last time step is $r_{u_{t-1}}$. The probability of requirement completion is estimated by the formula \eqref{goalcom} and the current requirement is predicted by maximizing the following probability:

\begin{equation}
\arg \max _{r_{u_t}} P\left( r_{u_t}\mid u_t, r_{u_{t-1}}\right)
\label{goalpre}
\end{equation}
\begin{equation}
P_{r_{u_t}}\left(l=1 \mid u_t, r_{u_{t-1}}\right)
\label{goalcom}
\end{equation}

\subsubsection{Encoder}
URef-(b) contains two encoders: ERNIE and Node2Vec. ERNIE is a large-scale pre-training model to extract semantic features in utterances. Transformer is used to encode text by ERNIE, which is a Seq2Seq model with the ability to encode variable-length sentences into fixed-length vectors based on Self-Attention. After entering the user text expression into ERNIE, the vector $v_s$ which represents the semantic information of utterance is obtained.

Node2Vec is a Graph Embedding model, based on random walk algorithm to extract requirement transition features. Node2Vec learns a mapping of nodes to a low-dimensional space of features by maximizing the likelihood of preserving network neighborhoods of nodes. Thus, the vector $v_n$, representation of each node in the backbone graph, can be obtained.

\subsubsection{Completion Network}
Compared with the requirement information $v_n$, the completion network prefer to use the utterance information $v_s$ when estimating whether the requirement is complete. Therefore, this network applies forget gate mechanism of LSTM in the calculation of a weaken factor $f_c$ which can control how many requirement information will be dropped out. This weaken factor $f_c$ is defined as:
\begin{equation}
f_{c}=\operatorname{sigmoid}\left(W_{c} v_{s} + U_{c} v_{n}+b_{c}\right)
\label{fc}
\end{equation}

In the formula \eqref{fc}, $W_{c}$ and $U_{c}$ are the weighted matrices and $b_c$ is the bias. In particular, node vector $v_n$ has the same dimension as sentence vector $v_s$.

After the requirement transition information forgotten mechanism, completion vector $v_c$ that participates in the estimation of requirement completion, which is defined as:
\begin{equation}
v_{c}=f_c \times v_s, f_c \in [0,1]
\label{vc}
\end{equation}

Then, completion vector $v_c$ which is utilized in the binary-class classification task, which is defined as:
\begin{equation}
y_{c}=\operatorname{softmax} \left( W_1 v_c + b_1 \right)
\label{twoclass}
\end{equation}

\subsubsection{Prediction Network}
In the requirement prediction, both the utterance information $v_s$ and the requirement information $v_n$ are applied. Firstly, the prediction network calculates a reinforce factor $f_r$ which can control how many requirement transition information will be fused. This reinforce factor $f_r$ is calculated in the same way as $f_c$:

\begin{equation}
f_{r}=\operatorname{sigmoid}\left(W_{r} v_{s} + U_{r} v_{n}+b_{r}\right) \in [0,1]
\label{fr}
\end{equation}

Then, the weakened vector $v_c$ output from the completion network is merged with enhanced sentence vector as:

\begin{equation}
f_{p}=\operatorname{tanh}\left(W_{p} v_{s} + U_{p}(f_{r} v_{c}) +b_{p}\right) \in [-1,1]
\label{fp}
\end{equation}

The reset gate mechanism (similar to GRU) applies in the calculation of the fusion factor $f_p$, which can not only drop out the information in $v_n$ that has nothing to do with the requirement prediction, but also can reinforce the information related to the prediction in $v_s$.

Finally, the fusion vector $v_p$ is defined in the formula \eqref{vp}, which is utilized to predict the requirement of the current utterance as the formula \eqref{multiclass}.

\begin{equation}
v_{p}=f_{p} v_{n} + v_s) +b_{p}
\label{vp}
\end{equation}
\begin{equation}
y_{p}=\operatorname{softmax} \left( W_2 v_p + b_2 \right)
\label{multiclass}
\end{equation}

\section{Recommend Response Generation} 
\label{rrg}

\begin{figure*}[!t]
	\centering
	\includegraphics[width=0.7\linewidth]{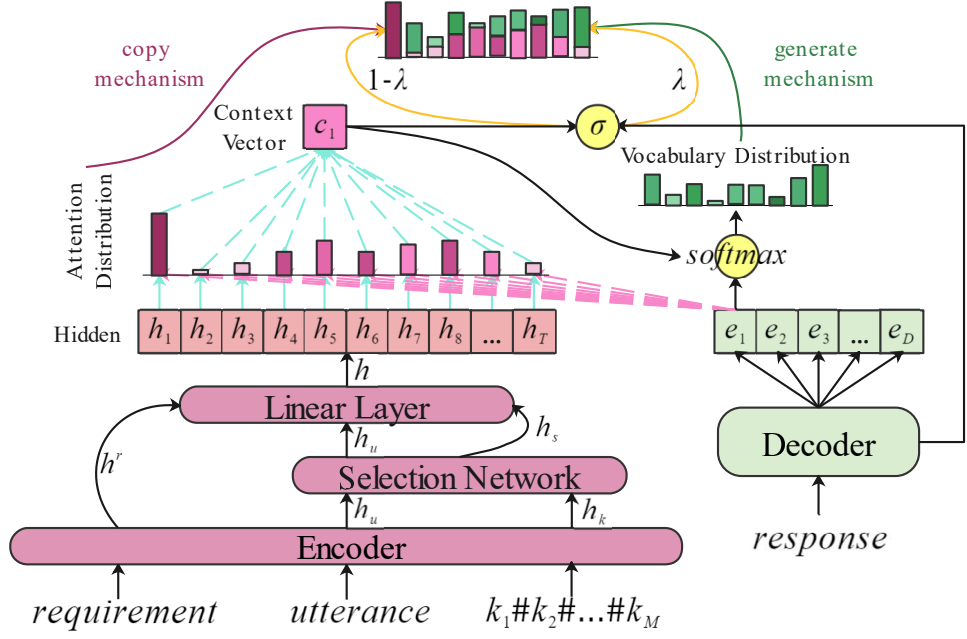}
	\caption{The architecture of SaRSNet, which is basically an encoder-decoder model.}
	\label{SaRSNet_figure}
\end{figure*}

\subsection{Overview of The Model}
As shown in Fig.~\ref{SaRSNet_figure}, SaRSNet is basically an encoder-decoder model to deal with text generation task. The requirement node, utterance and service resources will be taken as input by the same Encoder. The requirement node is a phase predicted by URef, which represents a kind of user's requirement. The utterance is current user's expression text. The service resources is a set of SPO triples in Personal KB, and their domain is consistent with the domain of requirement node after filtering by Algorithm 3. The service resource most relevant to the utterance is selected based on Selection Network. The structure of Decoder in SaRSNet is the same as the Encoder. In order to enable SaRSNet to replicate the selected service resource during decoding, there is a $\lambda$ to control the switching between copy mechanism and generation mechanism. Specifically, $\lambda$ is used as a soft switch to choose between generating a word from the vocabulary by sampling from the Vocabulary Distribution, or copying a word from the input sequence (contains service resource) by sampling from Attention Distribution. 

\subsection{Resource Selection}
\begin{figure}[!ht]
	\centering
	\includegraphics[width=0.8\linewidth]{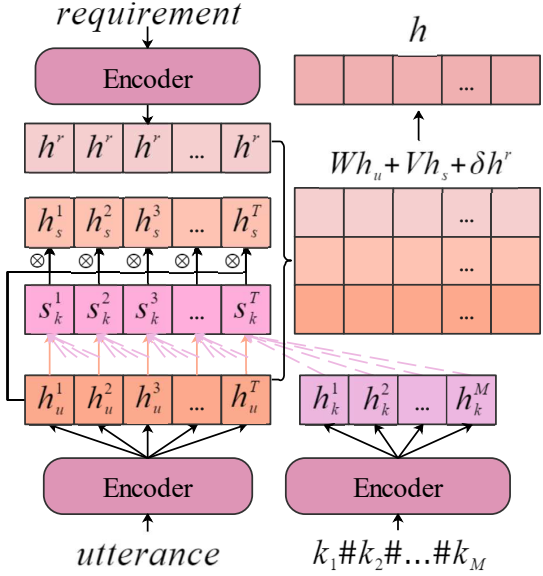}
	\caption{The details of Seclection Network in SaRSNet.}
	\label{seclection_network}
\end{figure}

During the encoding, Selection Network choose the best service resource according to the user's utterance. The details of Selection Network is shown in Fig.~\ref{seclection_network}. The tokens of each input text are fed one-by-one into the Encoder (a single-layer bidirectional LSTM), producing a sequence of encoder hidden states $h_i$. Input the requirement, the Encoder's outputs $h^{r}$, which is the last hidden state. Input the utterance, the Encoder outputs $h_{u}$, which is composed of all hidden states $(h_{u}^{1}, h_{u}^{2},...,h_{u}^{T})$. $T$ represents the numbers of tokens in the utterance. The service resources are spliced into a sentence as the input of the Encoder, and \textless \#\textgreater is added at the end of each triplet. Then, the Encoder outputs $h_k$, which is composed of all hidden states $(h_{k}^{1}, h_{k}^{2},...,h_{k}^{M})$ of \textless \#\textgreater. $M$ represents the numbers of service resources. At the time step $t$, the score of $h_{u}^{t}$ on the $i-th$ service resource $h_{k}^{i}$ is calculated as:
\begin{equation}
s_{k}^{t i}=\frac{\exp \left(h_{u}^{t} h_{k}^{i}\right)}{\sum_{i=1}^{M} \exp \left(h_{u}^{t} h_{k}^{i}\right)}, i=1,2, \ldots, M
\label{each_score}
\end{equation}

The final score of $h_{u}^{t}$ on all service resources is defined as:

\begin{equation}
s_{k}^{t}=\mathrm{max} \left\{s_{k}^{t i}, i=1,2, \ldots, M\right\}
\label{final_score}
\end{equation}

Then, $h_{u}^{t}$ is scored by $s_{k}^{t}$ and Selection Network outputs $h_{s}^{t}$:

\begin{equation}
h_{s}^{t}=h_{u}^{t} s_{k}^{t}
\label{scoreing}
\end{equation}

Finally, $h_{u}$, $h_{s}$ and $h_{r}$ are compressed into final hidden states $h$ by formula \eqref{final_h}, where $h=(h_{1}, h_{2}, ..., h_{T})$

\begin{equation}
h=W h_{u}+V h_{s}+\delta h^{r} 
\label{final_h}
\end{equation}

\subsection{Response Generation}
After the resource selection during encoding, the Decoder of SaRSNet takes the final hidden states $h$ as input. In order to copy a word from the selected resource at time step $i$, the Attention Distribution $a_{it}$ is calculated as formula \eqref{Attention_Distribution}, where $t$ represents the time step of the Encoder and $e_i$ represents the hidden state of the Decoder at time step $i$. The time steps of the Decoder and Encoder are not consistent.

\begin{equation}
a_{i t}=\frac{\exp \left(e_{i-1} h_{t}\right)}{\sum_{k=1}^{M} \exp \left(e_{i-1} h_{k}\right)}, t=1,2, \ldots, M
\label{Attention_Distribution}
\end{equation}

Then, the context vector $c_i$ at time step $i$ is calculated as:
\begin{equation}
c_{i}=\sum_{t=1}^{T} \exp \left(a_{it} h_{t}\right)
\end{equation}

$c_i$ can be seen as a fixed-size representation of what has been read from the source for this step. The vocabulary distribution of generated words at time step $i$ is calculated as formula \eqref{vocabulary_Distribution}, where $V^{\prime}$, $V$, $b^{\prime}$ and $b$ are all learnable parameters.

\begin{equation}
P_{\text {vocab }}=\operatorname{softmax} \left(V^{\prime}\left(V\left[c_{i}, e_{i}\right]+b\right)+b^{\prime}\right)
\label{vocabulary_Distribution}
\end{equation}

SaRSNet utilize $\lambda$ as soft switch to choose between copy mechanism and generation mechanism at time step $i$, which is calculated as formula \eqref{soft_switch}. 

\begin{equation}
\lambda=\operatorname{sigmoid}\left(W_{\lambda}\left[c_{i}, e_{i}\right]+b_{\lambda}\right)
\label{soft_switch}
\end{equation}

Thus, the distribution of Decoder's output words at the time step $i$ is calculated as formula \eqref{soft_switch}.

\begin{equation}
y_{i}=\lambda P_{\text {vocab }}+(1-\lambda) \sum_{t=1}^{T} a_{i t}
\label{output_distribution}
\end{equation}

\renewcommand{\algorithmicrequire}{\textbf{Input:}}
\renewcommand{\algorithmicensure}{\textbf{Output:}}
\begin{algorithm}[hbt]
	\caption{Resource Classification Process}
	\label{al3}
	\begin{algorithmic}[1]
		\REQUIRE The set of domains, $\{D\}$; predicate in a SPO triple, $p$; similarity threshold $\theta$, set as 0.7
		\ENSURE The domain to which the current triple belongs, $d$, $d \in \{D\}$;
		\FOR{ $d_i$ in $\{D\}$}
		\STATE $ s = sim(t, d_i) $ \COMMENT{compute semantic similarity between $t$ and $d_i$ by SimNet}
		\IF{ $ s \geq \theta $ } 
		\STATE $ d = d_i$  
		\ENDIF
		\ENDFOR
		\RETURN $d$
	\end{algorithmic}
\end{algorithm}

\section{Experiments} \label{Experiments}

\subsection{Datasets} 
The experiments involve an open dataset named \textit{DuRecDial} (about 10k dialogues, 156k utterances), which contains user profile for personalized recommendation, relevant Personal KB (service resource) and clear requirements to meet during each conversation. \textit{DuRecDial} is split into train/dev/test data by randomly sampling 70\%/10\%/20\% data. In \textit{DuRecDial}, there are 20 requirements and seven domains of user profile in total. The mapping relation between these domains and user requirements is shown in Table.~\ref{sevendomains}. However, the service resources (triplets of the Personal KB) in \textit{DuRecDial} are not annotated with domains of user profile. The process of resource classification is shown in Algorithm \ref{al3}.

\begin{table}[htbp]
\caption{Domains of Requirements in User Profile \& Personal KB}
\label{sevendomains}
\begin{center}
    \begin{tabular}{ll}
    \hline
    \makecell[c]{Domain} & \makecell[c]{Requirments}  \\
    \hline
    \makecell[c]{Star} & \makecell[c]{chitchat about celebrities}  \\
    \makecell[c]{Movie} & \makecell[c]{recommend movie, ask movie name, ask starring role} \\
    \makecell[c]{Music} & \makecell[c]{recommend music, play music, music order, \\ ask music name} \\
    \makecell[c]{Food} & \makecell[c]{recommend food} \\
    \makecell[c]{POI} & \makecell[c]{recommend poi} \\
    \makecell[c]{News} & \makecell[c]{recommend news, news order, ask news type} \\
    \makecell[c]{Weather} & \makecell[c]{ask the weather, ask time, ask the date,\\ weather information push} \\
    \makecell[c]{$\ast$} & \makecell[c]{daily greetings, goodbye} \\
    \hline
    \end{tabular}
    \begin{tablenotes}
		\scriptsize
		\item $\ast$ denotes requirements that have nothing to do with user profile and personal KB.
	\end{tablenotes}
\end{center}
\end{table}

\subsection{Comparison Models}

For Sequence Planning, this paper compares the machine learning model SP-MLP with the traditional strategy in Algorithm \ref{al2}. To show the effectiveness of the strategy proposed in URef-(a), this paper compares two different inputs of \textit{SP-MLP}: original feature (OF) and additional feature (AF). OF is a one-hot encoded feature of user profile and personal KB given in the dataset. After concatenating two $1\times20$ vectors, OF becomes AF. These two vectors encode the \textit{preference satisfaction} and \textit{knowledge abundance} features of user requirement, These two features are proposed in \textit{Algorithm \ref{al2}}.

To show the advantages of URef-(b) for Real-time Detection, this paper compares it with the following text classification models:
\begin{itemize}
    \item[-] TextCNN: a typical deep learning algorithm for sentence classification tasks based on convolutional neural network\cite{TextCNN}.
    \item[-] LSTM: a special kind of RNN, capable of learning long-term dependencies in sentence\cite{LSTM}.
    \item[-] BiLSTM: bi-directional LSTM, can better capture the two-way semantic dependence in sentence.
    \item[-] BERT: a pre-trained model can be fine-tuned with just one additional output layer to obtain state-of-the-art results on a wide range of natural language processing tasks\cite{bert}.
    \item[-] ERNIE: a novel pre-trained language representation model enhanced by knowledge, achieving new state-of-the-art results on Chinese natural language processing tasks\cite{ernie}.
    \item[-] PaddlePALM: a fast, flexible, extensible and easy-to-use NLP large-scale pre-training and multi-task learning framework based on ERNIE\cite{Dnet}.
\end{itemize}

To show the advantages of SaRSNet for Recommend Response Generation, this paper compares it with the following text generation models:
\begin{itemize}
    % \item[-] S2S: a vanilla seq2seq model based on LSTM \cite{S2S}.
    \item[-] S2S-Attn: a model that introduces the attention mechanism into S2S \cite{S2S-Attn}, where S2S is a vanilla seq2seq model based on LSTM \cite{S2S}.
    \item[-] Pointer-Generator: a hybrid pointer-generator network that can copy words from the source text via pointing\cite{pointer}.
    \item[-] MGCG\_G: a multi-goal driven conversation generation framework, consists of five components: a Utterance Encoder, a Knowledge Encoder, a Goal Encoder, a Knowledge Selector, and a Decoder\cite{baseline}. 
\end{itemize}

\subsection{Evaluation Metrics}
Totally, six evaluation metrics were used in the experiments: Exact Match (EM), BLEU (BLEU-2)\cite{bleu2}, Averaged Goal Recall (AGR)\cite{agr}, Accuracy, Precision, Recall perplexity (PPL)\cite{ppl} and DISTINCT (DIST-2)\cite{dist2}.

For the task of sequence planning, EM and BLEU-2 are utilized to indicate the general performance of URef-(a) on sequence-level. EM indicates the percentage of predicted requirement sequence that match the ground truth exactly, which is often used to evaluate Constituency Parsing task. BLEU-2 measures the similarity between predicted sequence and real sequence, which is often used to evaluate Machine Translation task. Following the setting in previous work (\cite{agr}), this paper also measures the node-level performance of URef-(a) using AGR\footnote{In this paper, AGR calculates the average recall of all requirement nodes instead of the dialogue goal.}.

For the task of Requirement Completion Estimation and Current Requirement Prediction, this paper both uses Accuracy to measure the performance of URef-(b), which is consistent with the baseline. Moreover, F1 is used to measure the overall performance of the model additionally, which is the harmonic mean between the Recall and Precision.

For the task of Resource Selection, precision, recall and F1 scores are evaluated the performance of SaRSNet. When calculating the knowledge precision/recall/F1, this paper compares the generated results with the correct knowledge.

For the task of Response Generation , BLEU-2, PPL and DIST-2 are used to measure the performance of SaRSNet. These three common metrics quantify the relevance, fluency, and diversity of generated responses.

\subsection{Parameter Settings}
In the experiments, the Adam Optimizer\cite{adam} is used to train SP-MLP and the initial learning rate is set as $1e$-3. 

In URef-(b), $max\_seq\_len$ is set as 128. Adam Optimizer is also used and the initial learning rate is set to $2e$-5. Optional parameter $warmup\_proportion$ is set as 0.1 and $weight\_decay$ is set as 0.01. For requirement transition graph embedding learning, the window size $h$ of random walk is set as 3, the return value is set as 1, the $in\-out$ value is set as 1 and the dimension of graph-based requirement node embedding is set as 100.

\begin{figure*}[hbt]
    \centering                                              
    \subfigure[Exact Match]{%              
        \begin{minipage}[t]{0.65\columnwidth}
        \centering                                                 
        \includegraphics[width=\columnwidth]{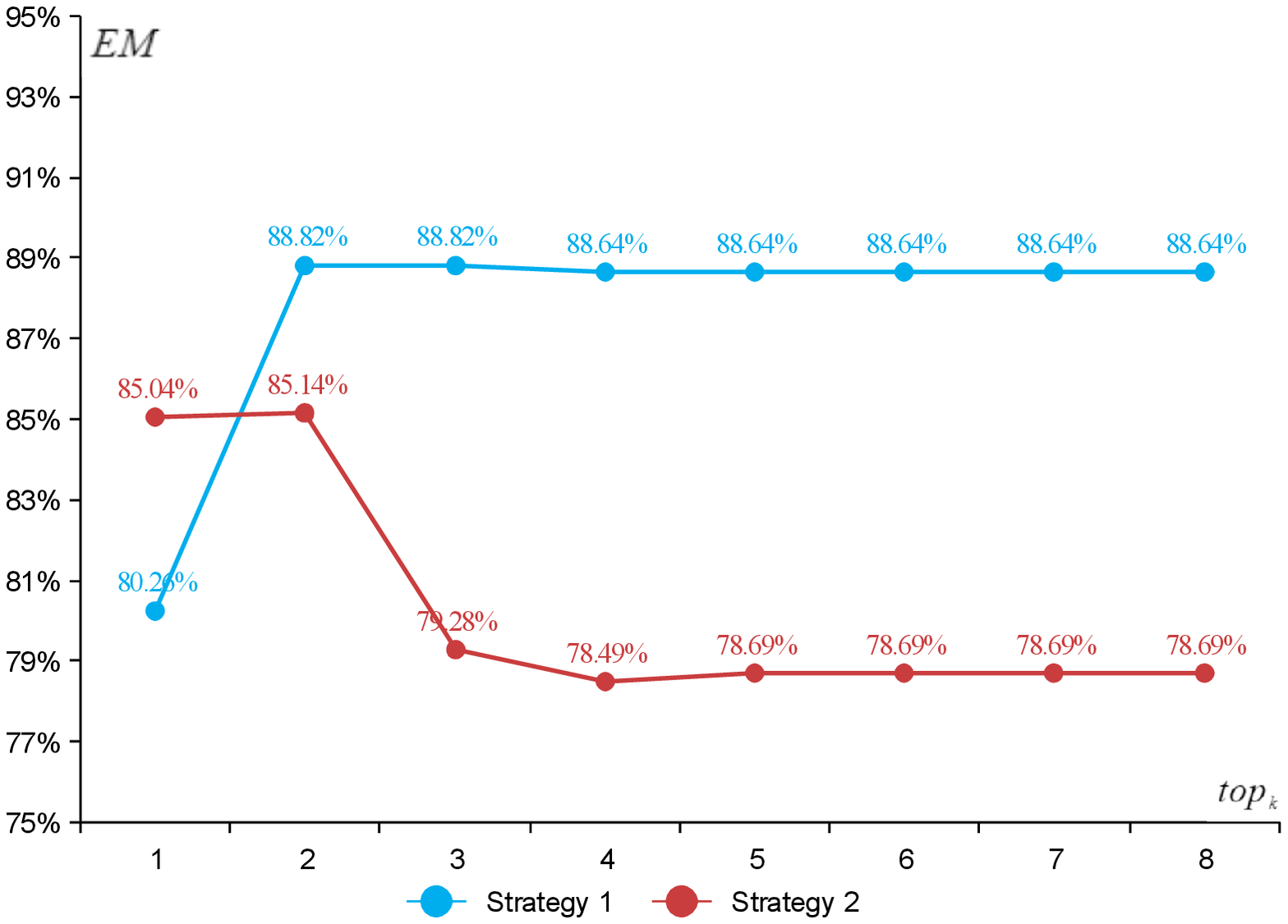} 
        \end{minipage}
        % \label{figa}
    }
    \subfigure[Averaged Goal Recall]{%              
        \begin{minipage}[t]{0.65\columnwidth}
        \centering                                                 
        \includegraphics[width=\columnwidth]{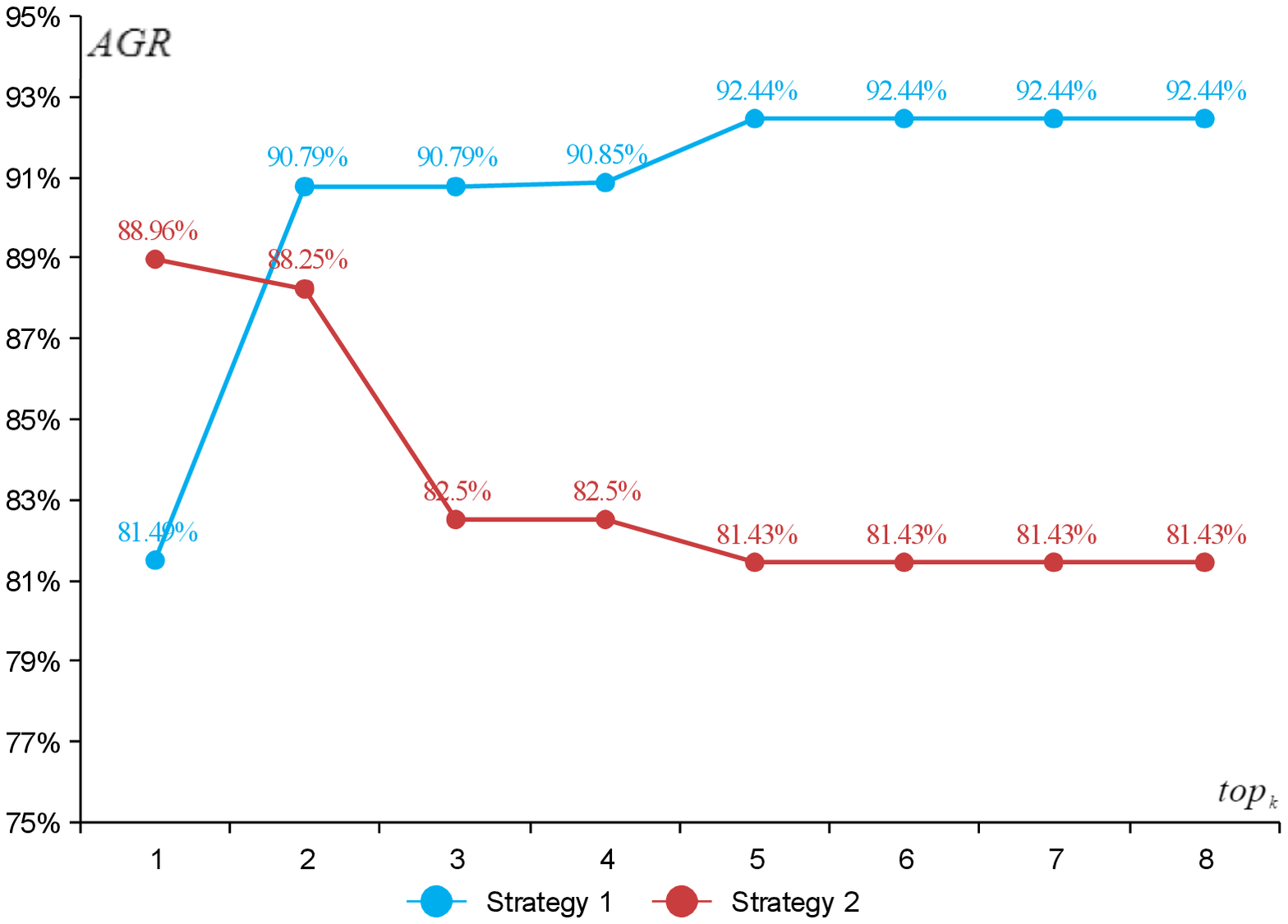}            
        \end{minipage}
        % \label{figb}
    }
    \subfigure[BLEU-2]{%          
        \begin{minipage}[t]{0.65\columnwidth}
        \centering                                                
        \includegraphics[width=\columnwidth]{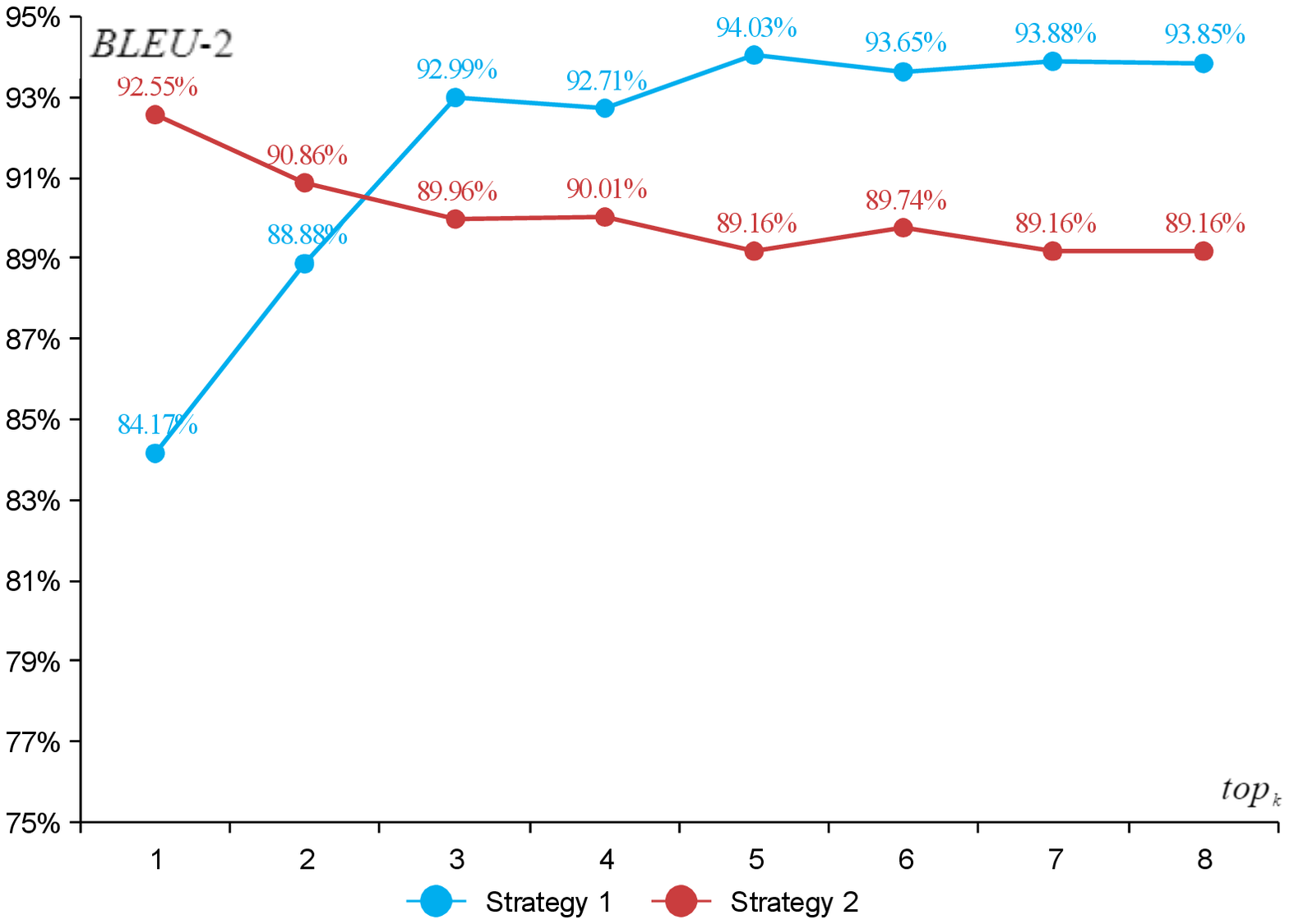}          
        \end{minipage}
        % \label{figc}
    }
    \caption{The performance of Algorithm 2 on different strategies}
    \label{strategy}                   
\end{figure*}

In SaRSNet, $max\_seq\_len$ is set as 256, $hidden\_size$ of Encoder is set as 512 and $dropout$ is set as 0.2. Adam Optimizer is utilized and the initial learning rate is set to $2e-4$. $warmup\_proportion$ is set to 0.2 and $weight\_decay$ is set as 0.01. The beam search is utilized during decoding and the $beam\_size$ is set as 10.

\section{Results} \label{Results}
\subsection{Sequence Planning}

\begin{figure*}[hbt]
    \centering                                           
    \subfigure[Exact Match]{%              
        \begin{minipage}[t]{0.65\columnwidth}
        \centering                                                 
        \includegraphics[width=\columnwidth]{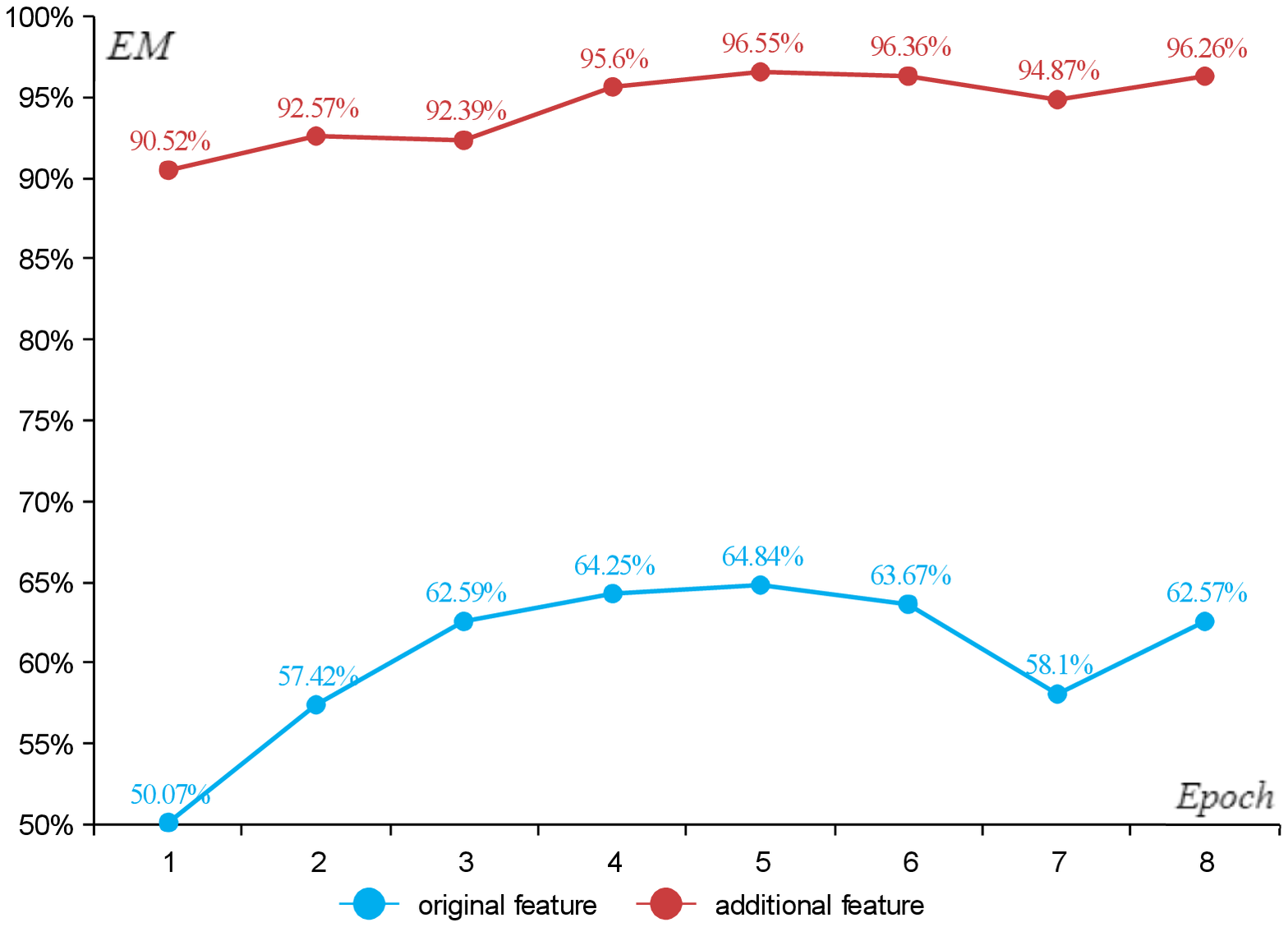} 
        \end{minipage}
    }
    \subfigure[Averaged Goal Recall]{%              
        \begin{minipage}[t]{0.65\columnwidth}
        \centering                                                 
        \includegraphics[width=\columnwidth]{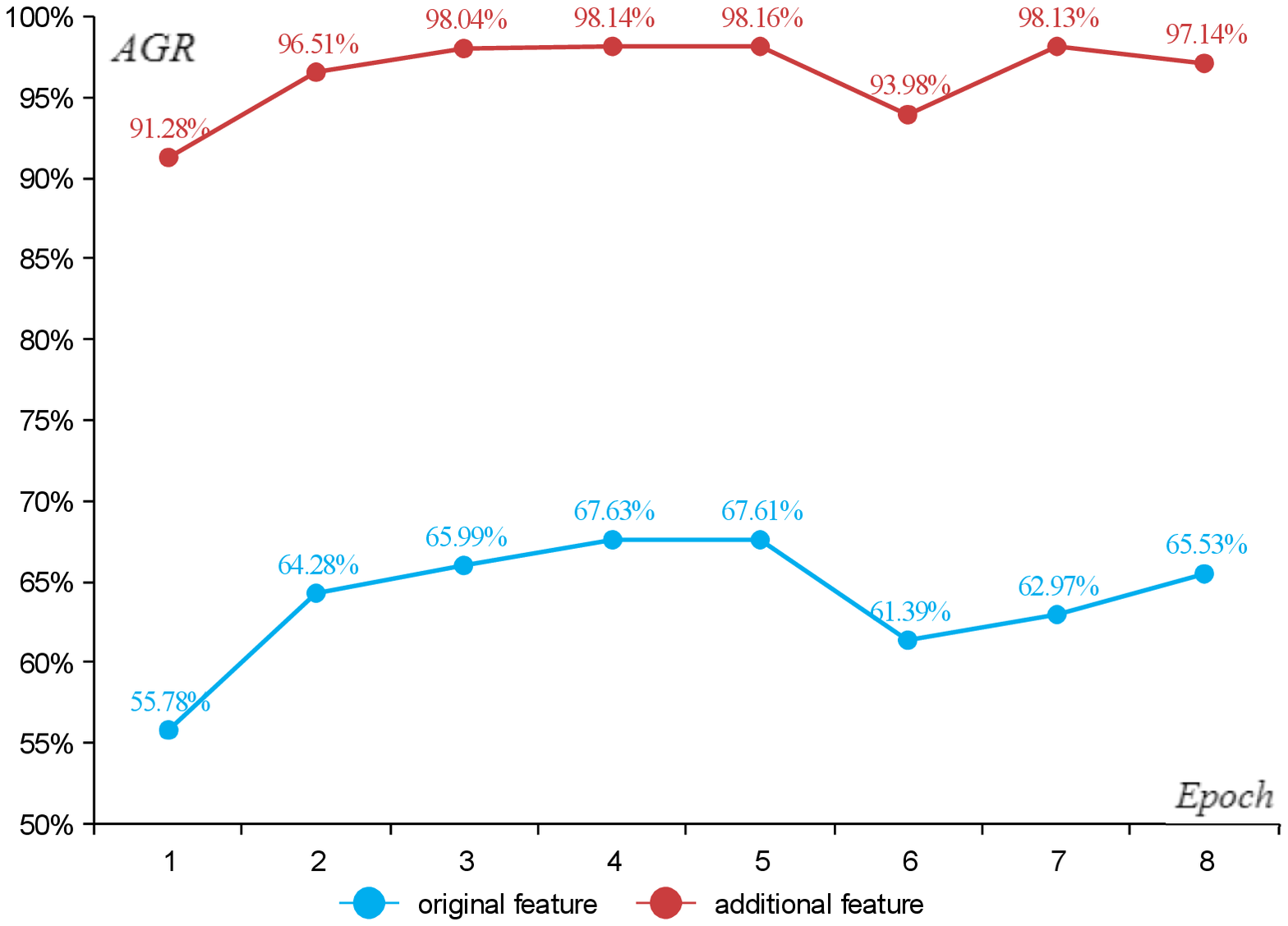}            
        \end{minipage}
    }
    \subfigure[BLEU-2]{%          
        \begin{minipage}[t]{0.65\columnwidth}
        \centering                                          
        \includegraphics[width=\columnwidth]{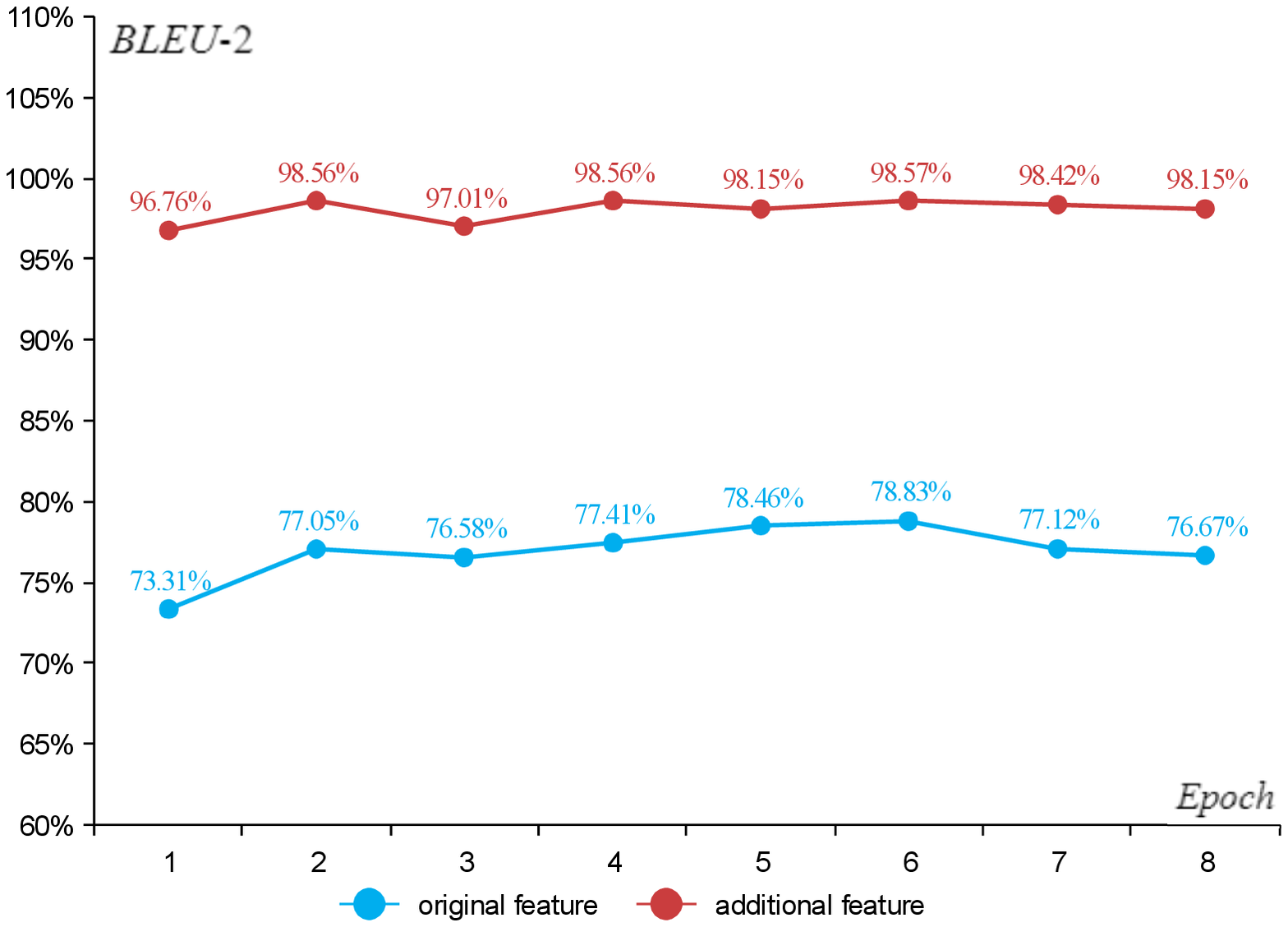}  
        \end{minipage}
    }
    \caption{The performance of SP-MLP on different features}
    \label{mlp_performance}                             
\end{figure*}

The proposed strategy is divided into two filters. The first filter selects $top_k$ sequences from the entire candidate sequences, and the second filter selects one of these $top_k$ sequences as the final result. The performance of the two strategies proposed in URef-(a) is shown in Fig.~\ref{strategy}. The size of $top_k$ represents the first filter's hardness, which means the diversity of the sequence set after screening. The larger the $top_k$, the easier the sequence in candidate set is to be selected. When $top_k$ exceeds a threshold, the first screening becomes meaningless. In order to choose a suitable $top_k$, this paper sets $top_k$ from 1 to 8 and conduct experiments respectively. In Fig.~\ref{strategy}, EM\textbf{1} and AGR\textbf{1} represent Exact Match and Averaged Goal Recall of Strategy \textbf{1} respectively. 

The results show that with the increase of $top_k$, the accuracy and BLEU-2 of Strategy 1 gradually increase, and both reach the maximum when $top_k=5$. Strategy 1 outperforms Strategy 2 overall. Due to the different precision of indicators, BLEU-2, AGR and EM decrease in order on Strategy1\&2. In the real world, the smarter strategy for each one in dialogue is to ``prescribe the right medicine", that is, first decide which topics (requirements) to talk about according to the other's preferred domain, and then screen the topics according to their own domain knowledge (service resource). The experimental results are in line with this reality.

However, with the increase of $top_k$, the performance of Strategy 2 gradually decreases. The explanation for this result is that in dialogue, based solely on the knowledge you have mastered, there will be a phenomenon of ``self-talking". If this strategy is adopted, the requirement nodes set by bot will deviate from user's interests. The larger the $top_k$, the more beneficial it is to oneself rather than the other party, and it will deepen the degree of divergence. \textit{preference satisfaction} and \textit{knowledge abundance} are also used for sequence planning, respectively, with unpleasant experimental results, as shown in Table.~\ref{goalplanning}. This proves the effectiveness of the proposed strategy.

\begin{table}[!h]
\caption{Requirement Sequence Planning Performance}
\label{goalplanning}
\begin{center}
    \begin{tabular}{llll}
    \hline
    \makecell[c]{Method} & \makecell[c]{EM} &\makecell[c]{AGR} &\makecell[c]{BLEU-2}  \\
    \hline
    \makecell[c]{Strategy1} & \makecell[c]{88.82} &\makecell[c]{92.44} &\makecell[c]{94.03}  \\
    \makecell[c]{Strategy2} & \makecell[c]{85.14} &\makecell[c]{88.96} &\makecell[c]{92.55}  \\
    \makecell[c]{Performance-only} & \makecell[c]{78.65} &\makecell[c]{81.43} &\makecell[c]{83.24}   \\
    \makecell[c]{Knowledge-only} & \makecell[c]{75.48} &\makecell[c]{78.91} &\makecell[c]{81.68}   \\
    \makecell[c]{SP-MLP-OF} & \makecell[c]{64.84} &\makecell[c]{67.63} &\makecell[c]{78.83}   \\
    \makecell[c]{SP-MLP-AF} & \makecell[c]{\textbf{96.55}} &\makecell[c]{\textbf{98.16}} &\makecell[c]{\textbf{98.57}}   \\
    \hline
    \end{tabular}
\end{center}
\end{table}

Besides, this paper compares two different inputs of SP-MLP. The experimental results are shown in Fig.~\ref{mlp_performance}. With the increase of the epochs, the performance of OF and AF gradually increases, and it drops when $epoch=$6 or 7. Although OF's performance is much lower than Strategy1\&2, AF's performance has been greatly improved compared with OF. EM of AF is the highest when $epoch=5$, reaching 96.55\%. This further proves the effectiveness of the strategy.

The best experimental results of all the models and strategies are shown in Table.~\ref{goalplanning}. The performance of SP-MLP-OF is the worst, followed by the performance of using \textit{knowledge abundance} and \textit{preference satisfaction} separately. SP-MLP-AF has the best performance, which is 31.71\%, 30.53\% and 19.74\% higher than OF in EM, AGR and BLEU-2 respectively. Compared with the best Strategy 1, AF has improved EM, AGR and BLEU-2 by 7.73\%, 5.72\% and 4.54\% respectively. The results show that there is a certain gap between the strategy learned by neural network based entirely on the historical data and the operating rules in real world. The features of \textit{knowledge abundance} and \textit{preference satisfaction} are input into SP-MLP, which reflects the strategy of human-machine dialogue and helps SP-MLP find the learning direction. Compared with the strategy algorithm, neural network has higher generalization ability and fault tolerance, so SP-MLP with additional feature performs better. However, the length of the sequence generated by SP-MLP is constant, less than or equal to 6. SP-MLP is not enough to deal with complex real scenes, because it's a supervised model.

\subsection{Real-time Detection}
Real-time Detection is divided into Requirement Completion Estimation and Current Requirement Prediction. As shown in Table.~\ref{goalclassify}, the former has fewer labels than the latter, so the accuracy is higher. To evaluate the contribution of the Encoder (ERNIE) in URef-(b), this paper conducts a comparative experiment between traditional text classification models (TextCNN,LSTM, BiLSTM) and pre-trained models (BERT, ERNIE). The results show that the pre-training model is better, and the accuracy of ERNIE is 0.38\% and 0.24\% higher than that of BERT in Requirement completion and Requirement prediction, respectively. PaddlePALM uses ERNIE to extract semantic features in utterance, and jointly learns these two sub-tasks in the way of hard parameter sharing. Compared with ERNIE, the accuracy of PaddlePALM has increased by 0.52\% and 0.21\% in Requirement Completion and Requirement Prediction, respectively. This proves that there is a potential correlation between these two sub-tasks, and higher performance of each sub-task can be obtained through joint learning. The joint model, URef-(b), can achieve the accuracy scores of 96.67\%, 94.13\% for Requirement Completion and Requirement Prediction, an increase of 1.99\%, 2.91\% over PaddlePALM respectively, and an increase of 2.54\%, 2.91\% over the baseline respectively. It proves the effectiveness of the sub-nets in URef-(b) after the fusion of requirements transition feature.

\begin{table}[!h]
\caption{Real-time Detection Performance}
\label{goalclassify}
\begin{center}
    \begin{tabular}{lll}
    \hline
    \makecell[c]{Method} & \makecell[c]{Requirement Completion\\Acc/F1} &\makecell[c]{Requirement Prediction\\Acc/F1}   \\
    \hline
    \makecell[c]{baseline} & \makecell[c]{94.13/93.28} &\makecell[c]{91.22/90.66}   \\
    \makecell[c]{TextCNN} & \makecell[c]{91.26/90.12} &\makecell[c]{91.17/90.08}   \\
    \makecell[c]{LSTM} & \makecell[c]{90.27/89.92} &\makecell[c]{91.89/90.15}   \\
    \makecell[c]{BiLSTM} & \makecell[c]{92.31/91.34} &\makecell[c]{91.65/90.26}   \\
    \makecell[c]{BERT} & \makecell[c]{93.78/91.94} &\makecell[c]{90.77/90.10}   \\
    \makecell[c]{ERNIE} & \makecell[c]{94.16/93.37} &\makecell[c]{91.01/90.71}   \\
    \makecell[c]{PaddlePALM} & \makecell[c]{94.68/93.39} &\makecell[c]{91.22/90.67}   \\
    \makecell[c]{URef-(b)} & \makecell[c]{\textbf{96.67/95.16}} &\makecell[c]{\textbf{94.13/93.62}}   \\
    \hline
    \end{tabular}
\end{center}
\end{table}

\begin{table*}[!t]
	\caption{The generated responses for different inputs}
	\label{tab_cs}
	\centering
	\begin{tabular}{ m{0.14\linewidth}< {\centering} m{0.45\linewidth}< {\centering} m{0.1\linewidth}<{\centering} m{0.16\linewidth}<{\centering}}
		\hline
		\makecell[c]{\textbf{Truth Response}} & \makecell[c]{\textbf{Generated Response}} & \makecell[c]{\textbf{Current}\\ \textbf{Requirement}} & \makecell[c]{\textbf{Service Resource}\\(Subject,Predicate,Object)}\\
		\hline
		
		\rowcolor{pink}
		\multicolumn{4}{l}{\textbf{User utterance}:Yes, idols are excellent. \textbf{Last Requirement}: chitchat about celebrities. \textbf{Completion}: Finished}\\
		
		\hline
		\multirowcell{4}{Would you like to \\ watch  the movie \\  \textcolor{blue}{Kung Fu Panda 3}? \\ This is a Hollywood \\ cartoon  with a \\ Chinese flavor.} & \makecell[c] {The movie \textcolor{blue}{Kung Fu Panda 3} is very good. I recommend it to you.} & \makecell[c]{recommend \\ movie} & (Jay Zhou, starring, \textcolor{blue}{Kung Fu Panda 3}) \\
		& Since you like him so much, do you want to watch his starring in \textcolor{blue}{Tian Court}? & recommend movie & \makecell[c]{-} \\
		& Jay Chou starred in \textcolor{blue}{Kung Fu Panda 3}. & \makecell[c]{-} & (Jay Zhou, starring, \textcolor{blue}{Kung Fu Panda 3}) \\
		& Yes, he's also a movie actor & \makecell[c]{-} & \makecell[c]{-}
		\\ 
		\hline
		
		\rowcolor{pink}
		\multicolumn{4}{l}{\textbf{User utterance}:Hello, any news about Jay Chou? \textbf{Last Requirement}: -. \textbf{Completion}: -.}\\
		
		\hline
% 		\multirowcell{4}{Hi, let's start the \\ news broadcast \\ immediately, Jay Chou  \\ released a new song \\ \textcolor{blue}{Waiting For You}, \\ which evoked \\ the memories \\ of youth in netizens}
		\multirowcell{4}{Hi, Jay Chou just \\ released a new song \\ \textcolor{blue}{Waiting For You}, \\ which evoked \\ the memories of \\ youth in netizens} & \makecell[c] {Then let me tell you a piece of news about Jay Chou, \\ he just released a new song \textcolor{blue}{Waiting For You}. } & news order & (Jay Zhou, releases, \textcolor{blue}{Waiting For You}) \\
		
		& Would you like to hear about Jay Chou's news? & news order & \makecell[c]{-} \\
		
		& Jay Chou released \textcolor{blue}{Waiting For You}, go and listen.  & \makecell[c]{-} & (Jay Zhou, releases, \textcolor{blue}{Waiting For You}) \\
		
		& Hello, there are. & \makecell[c]{-} & \makecell[c]{-} \\ 
		\hline
	\end{tabular}
	\begin{tablenotes}
		\scriptsize
		\item All sentences are in Chinese and they are translated into English in this table. Each pink row shows the user's utterance, the last user's requirement and the completion of the last requirement. Given one utterance, there are different responses generated with four types of input : ``+(-)r" and ``+(-)k". Below the pink row, the first line shows the output of SaRSNet when the current requirement and the service resource are taken as input. The remaining three lines show the output of MGCG\_G with different types of input respectively. ``-" represents the value of the item is none. The truth response is the real-valued reply to the utterance.  Moreover, the current requirement was predicted by URef and the service resource was selected by Selection Network in SaRSNet, both of which are taken as input.
	\end{tablenotes}
\end{table*}
% 是啊，偶像就是优秀啊 (关于明星的聊天, finished)
% 你要不要看一下功夫熊猫3这部电影？很有中国味儿的一部好莱坞动画片。
% 『 功夫熊猫3 』 这个 电影 很 好看 哦 ， 推荐 给 你"
% 既然你这么喜欢他 ，那他主演的[天庭外传]你要不要看一下？
% 周杰伦主演了功夫熊猫3
% 是的,他还是个电影演员.

% Jay Chou releases new song Waiting For You
% 新闻点播  哈喽, 有关于周杰伦的新闻吗?
% (周杰伦, 发布, 新歌《等你下课》)
% 你好， 马上 开始 新闻 播报 ， 周杰伦 发布 了 新歌 《 等 你 下课 》 ， 勾起 了 网友 们 对 青春 的 回忆 。

% 那 我 给 你 说 说 一个 关于 周杰伦 的 新闻 吧 。 truth:那有个周杰伦的新闻要不要听听呢 ？
% 周杰伦发布了等你下课,快去听听吧.

% 你好,有的.

\subsection{Response Generation}
The performance of models on Resource Selection and Response Generation are shown in Table.~\ref{reponse_generation}. All the models do not perform well with only user utterance as input on both two tasks, while improving when the requirement node is added to the input. This indicates the necessity of user requirements in the Recommend Response Generation. Moreover, "S2S-Attn +r.-k." performs worse than "S2S-Attn -r.+k.", because S2S-Attn does not have the ability to select service resource. Compared to S2S-Attn, Pointer-Generator is able to select resources based on copy mechanism, which is also introduced into SaRSNet. However, MGCG\_G performs better than Pointer-Generator due to its three encoders. This indicates the effectiveness of a strategy for encoding the three inputs (utterance, requirement, resources) separately, which is also introduced into SaRSNet. SaRSNet overall outperforms MGCG\_G on six metrics of two tasks, which shows that the modeling approach on resource selection and text generation is reasonable.
\begin{table}[!h]
\caption{Response Generation Performance}
\label{reponse_generation}
\begin{center}
    \begin{tabular}{lll}
    \hline
    \makecell[c]{Method} & \makecell[c]{Resource Selection \\Precision/Recall/F1} &\makecell[c]{Response Generation \\BLEU-2/DIST-2/PPL}   \\
    \hline
    \makecell[c]{S2S-Attn -r.-k.} & \makecell[c]{0.253/0.205/0.226} &\makecell[c]{0.152/0.018/25.82}   \\
    \makecell[c]{S2S-Attn +r.-k.} & \makecell[c]{0.265/0.219/0.24} &\makecell[c]{0.171/0.024/23.08}   \\
    \makecell[c]{S2S-Attn -r.+k.} & \makecell[c]{0.258/0.212/0.232} &\makecell[c]{0.173/0.021/22.85}   \\
    \makecell[c]{S2S-Attn +r.+k.} & \makecell[c]{0.274/0.235/0.253} &\makecell[c]{0.195/0.026/20.13}   \\
    \hline
    \makecell[c]{Pointer-Generator -r.-k.} & \makecell[c]{0.256/0.211/0.231} &\makecell[c]{0.169/0.035/18.57}   \\
    \makecell[c]{Pointer-Generator +r.-k.} & \makecell[c]{0.268/0.225/0.244} &\makecell[c]{0.177/0.034/18.11}   \\
    \makecell[c]{Pointer-Generator -r.+k.} & \makecell[c]{0.352/0.321/0.335} &\makecell[c]{0.181/0.044/17.82}   \\
    \makecell[c]{Pointer-Generator +r.+k.} & \makecell[c]{0.358/0.329/0.343} &\makecell[c]{0.195/0.059/17.24}   \\
    \hline
    \makecell[c]{MGCG\_G +r.+k.} & \makecell[c]{0.401/0.377/0.383} &\makecell[c]{0.219/0.052/17.69}   \\
    \makecell[c]{SaRSNet +r.+k.} & \makecell[c]{\textbf{0.408/0.388/0.398}} &\makecell[c]{\textbf{0.225/0.083/16.12}}   \\
    \hline
    \end{tabular}
    \begin{tablenotes}
		\scriptsize
		\item All models take user's utterance as original input. ``+(-)r." represents add (not add) the requirement node to the original input. ``+(-)k." represents add (not add) the service resource to the original input. For both ``S2S-Attn +r.+k." and ``Pointer-Generator +r.+k.", this paper concatenates the utterance, the requirement node, and the related service resources into a piece of text as input.
	\end{tablenotes}
\end{center}
\end{table}
\subsection{Case Study}
To reveal the impact of URef on SaRSNet in the proposed Two-phase Requirement Elicitation Method, the generated responses for different inputs are shown in Table.~\ref{tab_cs}. Each pink row shows the user's utterance, the last user's requirement and the completion of the last requirement. The current requirement was predicted by URef and the service resource was selected by Selection Network of SaRSNet. Given one utterance, there are different responses generated with four types of input : ``+r+k", ``+r-k", ``-r+k", and ``-r-k". ``+r+k" means take both current requirement and service resource as input.

\begin{CJK*}{UTF8}{gbsn}
	\begin{itemize}
		\item \textbf{Utterance 1:} Yse, idols are excellent? (是啊，偶像就是优秀啊)\\
		According to the predictions of URef-(a), the last user requirement is \textit{chitchat about celebrities}. Based on the utterance, URef-(b) supposes it has been finished and predicts the user's current potential requirement is \textit{recommend movie}. Thus, the bot should generate a response grounded on the current requirement and service resource to initiate a new conversation. Without the requirement and resource, the generated response (Yes, he's also a movie actor. 对的，他还是个演员) is security and hollow when only the utterance is taken as input. With the requirement, the anomaly generated response (Since you like him so much, do you want to watch his starring in Tian Court? 既然你这么喜欢他，那他主演的天庭外传你要不要看一下？) is more relevant to the topic. But the recommended item (Kung Fu Panda 3) is inconsistent with the truth response (Tian Court) when inputs is missing the service resource. In fact, Tian Court have nothing to do with Jay Chou. Lack of resources leads to factual errors in generated response. Although, the recommended item appears in the generated response (Jay Chou starred in Kung Fu Panda 3. 周杰伦主演了功夫熊猫3) with the service resource, the response is only a statement of fact and lacks the tone of recommendation cause of the missing of the current requirement. With the requirement and the resource, SaRSNet generates an appropriate response (The movie Kung Fu Panda 3 is very good. I recommend it to you.功夫熊猫3这部电影很叫座，我推荐给你看看) based on the utterance, which is closer to the truth response.
		
		\item \textbf{Utterance 2:} Hello, any news about Jay Chou? (哈喽, 有关于周杰伦的新闻吗?)\\
		Different from \textbf{Utterance 1}, \textbf{Utterance 2} is the first sentence initiated by the user in a conversation. Therefore, the last requirement does not exist. URef-(b) predicts the current requirement is \textit{news order} and the completion is unfinished. The generated response needs to answer the user's question. As in the case of \textbf{Utterance 1}, lack of the requirement and the resource will result in a security, hollow or anomaly response. Compared with others, the response (Then let me tell you a piece of news about Jay Chou, he just released a new song Waiting For You. 那我给你说一个关于周杰伦的新闻吧，他刚刚发布新歌等你下课。) generated by SaRSNet based on the requirement and the resource meets the requirement of the user well. This further demonstrates that both user requirement predicted by URef and service resources selected by SaRSNet are important. 
% 		The performance of URef deeply influence the performance of SaRSNet.
	\end{itemize}
\end{CJK*}

\section{Conclusion} \label{conclusion}
This paper proposed a novel user requirement elicitation framework (URef), which is divided into two parts: URef-(a) and URef-(b). Grounded on user’s preference and Personal KB, URef-(a) prepares the user's potential requirement chain before the conversation. Based on the user's utterance, URef-(b) judges whether the last requirement is met and then predicts user's requirement in the current-turn. When the requirement is changed, URef-(a) will re-plan a sequence with the new starting point. 

To deal with Recommend Response Generation task, a novel Seq2Seq Model (SaRSNet) is proposed. SaRSNet selects the right service resource based on scoring mechanism, then generates an appropriate response text to user grounded on user's utterance, selected service resource and user's requirement node.

However, user profile is known beforehand in this work, not obtained by bots through questioning. Thus, it is interesting to explore how to obtain the user profile during the conversation. This work will be left as the future work.

% \appendices

% % you can choose not to have a title for an appendix
% % if you want by leaving the argument blank
% \section{}

\section*{Acknowledgment}
This work is partially supported by the National Key Research and Development Program of China(2018YFB1402500), the National Science Foundation of China(61802089, 61772155, 61832004, 61832014).

\bibliographystyle{IEEEtran}
\bibliography{sample-base}
% 减小间距
\vspace{-12 mm} 

\begin{IEEEbiography}
[{\includegraphics[width=1in,height=1.25in,clip,keepaspectratio]{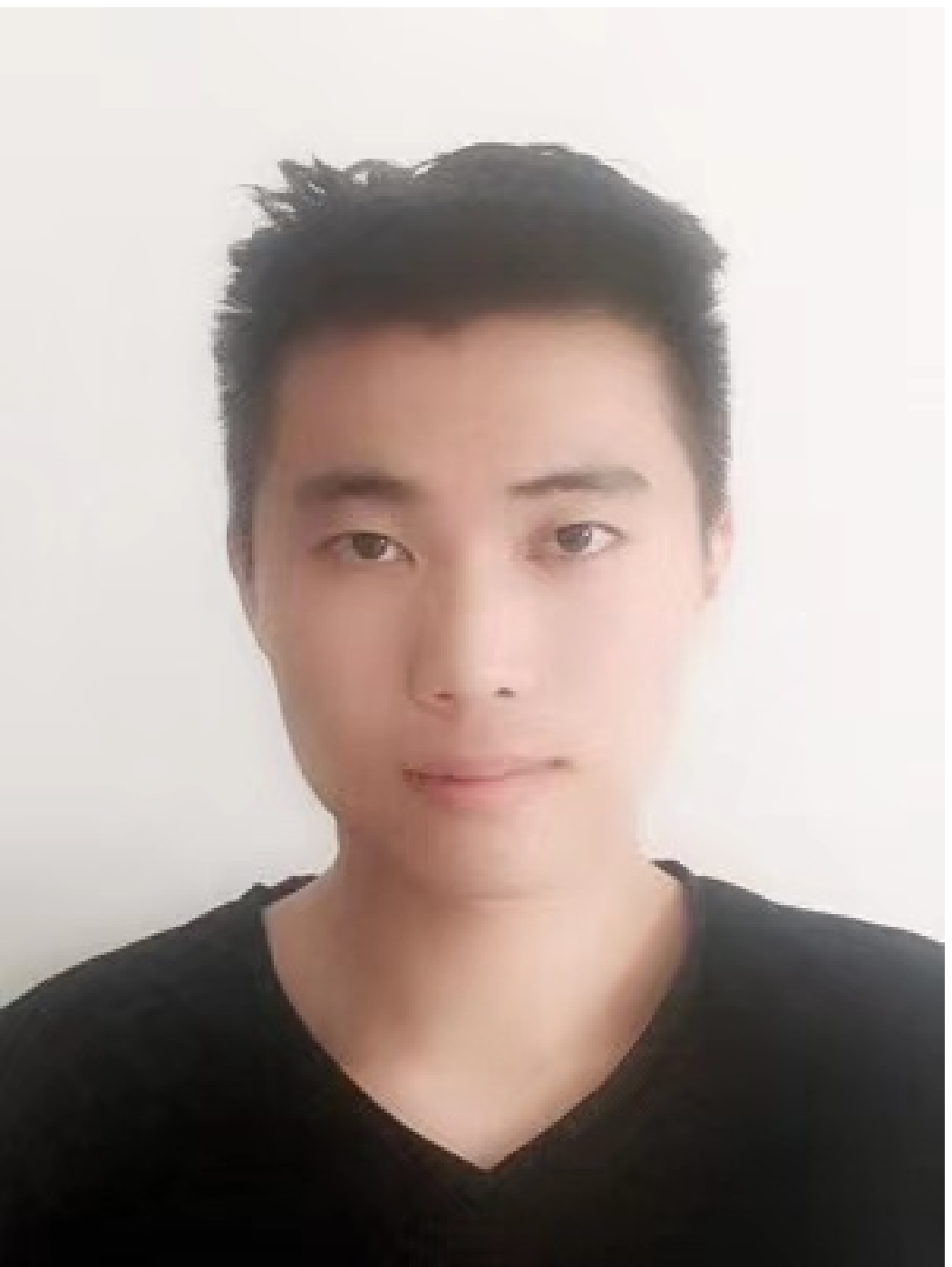}}]
{Bolin Zhang}
received the MS degree in Computer Science from the Harbin Institute of Technology (HIT), China. He is currently an Ph.D at the ICES Lab of HIT. His research interests include cognitive service, software service bot, and dialogue system.
\end{IEEEbiography}

\begin{IEEEbiography}
[{\includegraphics[width=1in,height=1.25in,clip,keepaspectratio]{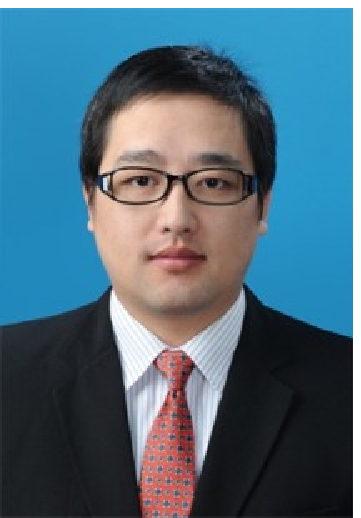}}]
{Zhiying Tu}
received the MS degree in Computer Science from the Harbin Institute of Technology (HIT), China. He received a Ph.D degree in Computer Integrated Manufacturing (Productique) from the University of Bordeaux, France. Since 2013, He began to work at HIT. His research interests include Service Computing, Enterprise Interoperability, and Cognitive Computing. He has 20 publications as edited books and proceedings, refereed book chapters,and refereed technical papers in journals and conferences. He is the member of IEEE Computer Society. 
\end{IEEEbiography}

\begin{IEEEbiography}
[{\includegraphics[width=1in,height=1.25in,clip,keepaspectratio]{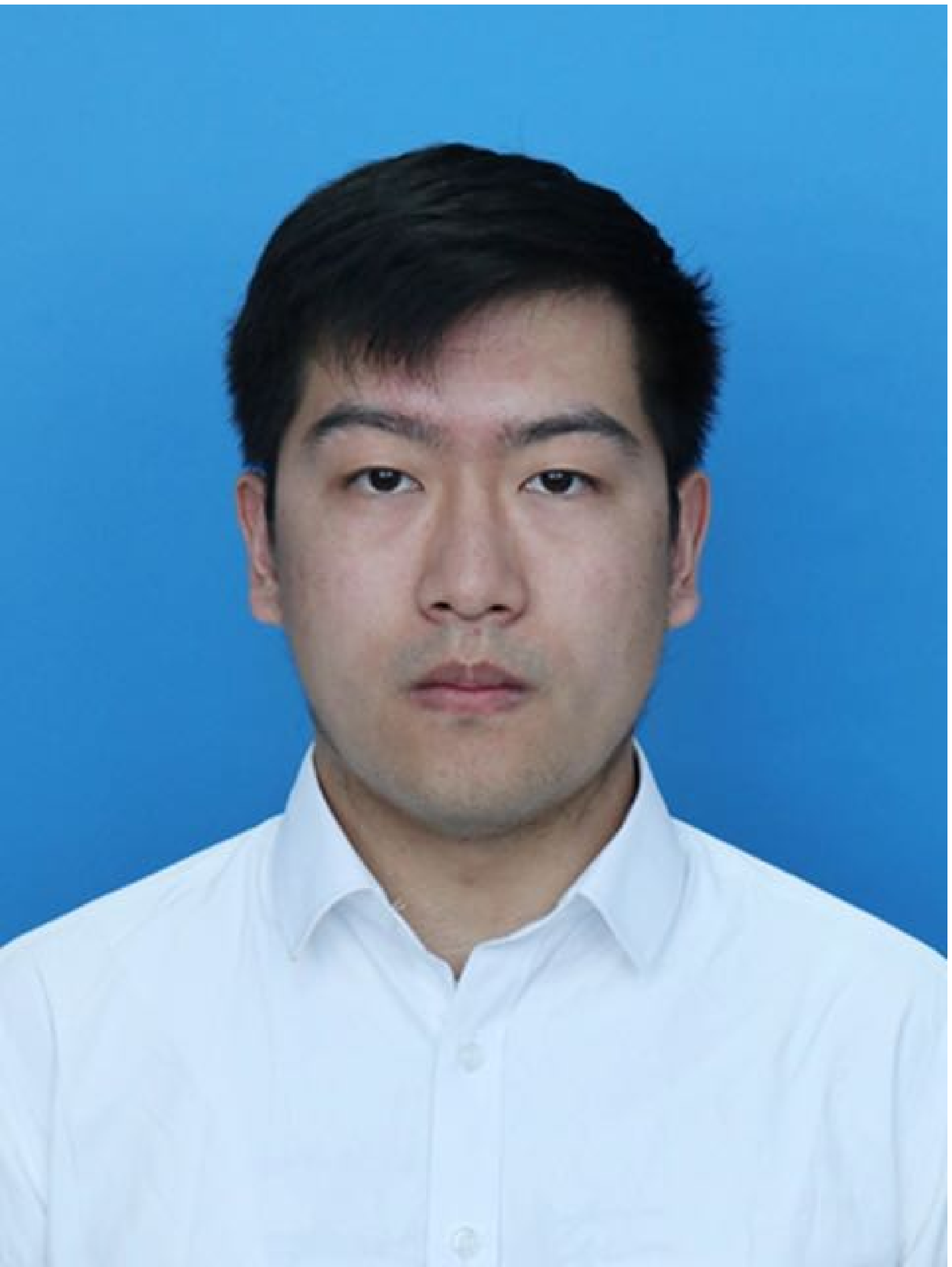}}]
{Yunzhe Xu}
received the BCs degree in Computer Science from the Harbin Institute of Technology (HIT), China. His research interests include Service Computing, text generation and dialogue system.
\end{IEEEbiography}
\vspace{-100 mm} 
\begin{IEEEbiography}
[{\includegraphics[width=1in,height=1.25in,clip,keepaspectratio]{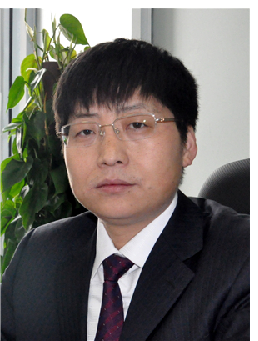}}]
{Dianhui Chu}
received the MS and PhD degrees in Computer Science from the Harbin Institute of Technology (HIT), China. He is a professor and dean at School of Computer Science and Technology in HIT. His research interests include Service Computing.
\end{IEEEbiography}

\vspace{-100 mm} 

\begin{IEEEbiography}
[{\includegraphics[width=1in,height=1.25in,clip,keepaspectratio]{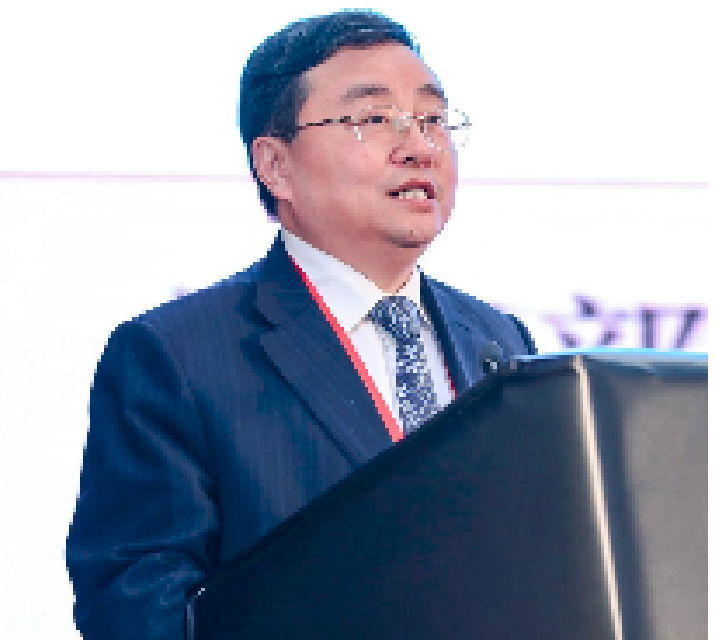}}]
{Xiaofei Xu}
received the MS and PhD degrees in Computer Science from the Harbin Institute of Technology (HIT), China. He is a professor at School of Computer Science and Technology, and vice president of HIT. His research interests include enterprise intelligent computing, services computing, Internet of services, and data mining. He is the associate chair of IFIP TC5 WG5.8, chair of INTEROP-VLab China Pole, fellow of China Computer Federation (CCF), and the vice director of the technical committee of service computing of CCF. He is the author of more than 300 publications. He is member of the IEEE and ACM. 
\end{IEEEbiography}

\end{document}